\documentclass[lettersize,journal,article]{IEEEtran}
\usepackage{amsmath,amsfonts}
\usepackage[linesnumbered,ruled]{algorithm2e}
\usepackage{arydshln}
\usepackage[caption=false,font=normalsize,labelfont=sf,textfont=sf]{subfig}
\usepackage{textcomp}
\usepackage{stfloats}
\usepackage{url}
\usepackage{verbatim}
\usepackage{graphicx}
\usepackage{cite}
\usepackage{orcidlink}
\usepackage{lipsum}
\usepackage[hang,flushmargin]{footmisc}
\newcommand\modelname{\textsc{Decider}}
\usepackage{amsmath, bm}
\usepackage{graphicx}
\usepackage{bbold}
\usepackage{multirow}
\usepackage{threeparttable}
\usepackage{caption} 
\usepackage{balance}
\usepackage{multirow}
\usepackage{booktabs}
\usepackage{hyperref}
\usepackage{dsfont}
\usepackage{enumitem}
\usepackage{arydshln}
\usepackage{bbding}
\usepackage{xcolor,colortbl}
\definecolor{gray}{gray}{0.93}
\definecolor{change}{RGB}{0,0,0}
\usepackage{arydshln}
\usepackage{tikz}
\newcommand*\circled[1]{\tikz[baseline=(char.base)]{
            \node[shape=circle,draw,inner sep=1pt] (char) {#1};}}

\definecolor{purple}{RGB}{112,48,160}
\definecolor{ocean}{RGB}{2,154,152}
\definecolor{blue}{RGB}{0,0,0}
\definecolor{red}{RGB}{192,0,0}
\definecolor{light}{RGB}{67, 104, 247}
\newcommand\pbot{\textsc{P}$^2$\textsc{BOT}}
\newcommand\neurologic{\textsc{NeuroLogic}}
\newcommand\csquare{Persona C$^2$}
\usepackage{tabularx,array}

\newcolumntype{C}{>{\centering\arraybackslash}X}
\newcommand{\tabcell}[1]{\begin{tabular}{@{}c@{}}#1\end{tabular}}
\begin{document}

\title{\modelname{}: A Dual-System Rule-Controllable Decoding Framework for Language Generation}

\author{\IEEEauthorblockN{
Chen Xu~\orcidlink{0000-0002-3495-4238},
Tian Lan~\orcidlink{0000-0002-5200-1537},
Yu Ji~\orcidlink{0009-0006-9306-6966},
Changlong Yu~\orcidlink{0000-0002-4758-9014},
Wei Wang~\orcidlink{0000-0002-2908-3060},
Jun Gao~\orcidlink{0000-0003-0152-9654},
\\
Qunxi Dong~\orcidlink{0000-0002-0484-3019}, 
Kun Qian~\orcidlink{0000-0002-1918-6453},
Senior Member, IEEE,
Piji Li~\orcidlink{0000-0003-1474-3692},
Wei Bi~\orcidlink{0000-000x-xxxx-xxxx},
 and 
Bin Hu\textsuperscript{\Envelope}~\orcidlink{0000-0003-3514-5413}, Fellow, IEEE}\\
\thanks{
Chen Xu, Yu Ji, Qunxi Dong, Kun Qian, Bin Hu are with the Key Laboratory of Brain Health Intelligent Evaluation and Intervention, Ministry of Education, School of Medical Technology, Bejing Institute of Technology (BIT), China (e-mail: chenxu05037@bit.edu.cn). Tian Lan is with the School of Computer Science and Technology, BIT, China. Piji Li is with Nanjing University of Aeronautics and Astronautics, Nanjing, China. 
}
\thanks{\Envelope \ Corresponding author: Bin Hu}}

\markboth{}%
{Shell \MakeLowercase{\textit{et al.}}: A Sample Article Using IEEEtran.cls for IEEE Journals}

%
%

\maketitle

\begin{abstract}
Constrained decoding approaches aim to control the meaning or style of text generated by the pre-trained large language models (LLMs or also PLMs) for various tasks at inference time. 
However, these methods often guide plausible continuations by greedily and explicitly selecting targets. Though fulfilling the task requirements, these methods may overlook certain general and natural logics that humans would implicitly follow towards such targets.
Inspired by cognitive dual-process theory, in this work, we propose a novel decoding framework \modelname{} where the base LLMs are equipped with a First-Order Logic (FOL) reasoner to express and evaluate the rules, along with a decision function that merges the outputs of both systems to guide the generation.
Unlike previous constrained decodings, \modelname{} transforms the encouragement of target-specific words into all words that satisfy several high-level rules, enabling us to programmatically integrate our logic into LLMs.
%
%
%
Experiments on CommonGen and PersonaChat demonstrate that \modelname{} effectively follows given FOL rules to guide LLMs in a more human-like and logic-controlled manner.
\end{abstract}

\vspace{-1em}
\section{Introduction}

\IEEEPARstart{C}{ontrollable} natural language generation \cite{keskar2019ctrl, Dathathri2020Plug, qian2022controllable} aims to generate natural sentences that satisfy specific conditions,
such as sentiment~\cite{ghosh2017affect,li2020hierarchical}, topic~\cite{xing2017topic,liao2020topic,liu2022graph,NEURIPS2022_871cae8f,su2022language} and keywords~\cite{lin-etal-2020-commongen,li2023automatic,lan2022complex}. This work is initiated by the latter, i.e., keyword-controlled generation, where language generation is \textit{lexically} or \textit{semantically} constrained by certain target words (the terms ``target words'' and ``constrained words'' are interchangeable). These words can be used to control the meaning or style of the generated text.

We selected two representative and typologically complementary tasks in constrained language generation: CommonGen~\cite{lin-etal-2020-commongen}, a lexically-constrained text generation, and PersonaChat~\cite{zhang2018personalizing}, a semantically-constrained dialogue generation task. CommonGen requires the generated sentence to \textit{exactly cover all} given target words. For example, given a concept set {\small {\tt \{enjoy, classroom, students\}}}, a plausible output would be ``The {\small{\tt students enjoy}} learning in the {\small{\tt classroom}}''. Personalized response generation (PersonaChat)~\cite{zhang2018personalizing} requires the model to generate responses that are \textit{semantically consistent} with specific keywords from a given persona description. For instance, when asked ``Hi, what are you doing?'', the one with the persona ``I have {\small$\texttt{pets}$}'' might answer ``I just got home from walking my {\small$\texttt{dog}$}''

\begin{figure*}[!t]
\centering
\includegraphics[width=\textwidth]{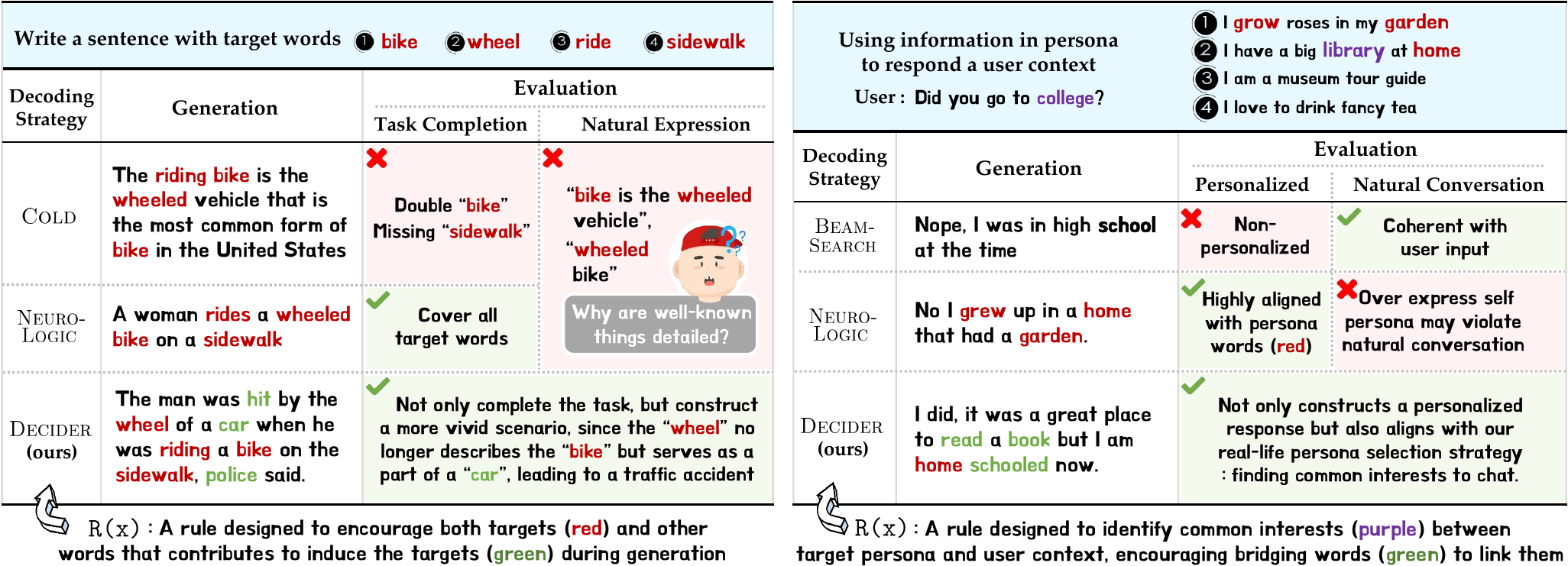}
\caption{
\textcolor{blue}{In the CommonGen (left) and PersonaChat (right) tasks, \modelname{} can employs and follows the rules in Table~\ref{tab:kb} to mitigate the greedy focus on targets, producing logic-controllable text that not only completes task requirements but also better aligns with natural human expression scenarios.}}
\label{fig:intro_case}
\vspace{-1em}
\end{figure*}

Recent research in this field can be divided into two categories.
The first category involves training (fine-tuning) a vanilla generative model~\cite{lin-etal-2020-commongen}, or designing more sophisticated model structures~\cite{fan-etal-2020-enhanced, majumder-etal-2020-like, wang2021neural, liu2021kg, carlsson2022fine} and generation processes~\cite{liu-etal-2020-impress, xu2022cosplay, wang2021contextualized, yang-etal-2023-bridging,wang-etal-2021-mention} that incorporate task-related prior knowledge. However, these methods often need to be fine-tuned or re-trained on large data due to the domain adaptation, or their customized model structure and generation process. 
%
%
With the popularity of pre-trained language models (PLMs), researchers are working on developing decoding methods that directly guide a pre-trained base model toward target keywords while keeping the text quality as much as possible without training~\cite{anderson-etal-2017-guided, hokamp-liu-2017-lexically, post-vilar-2018-fast, Dathathri2020Plug, pascual-etal-2021-plug-play, lu-etal-2021-neurologic}. Specifically, the representatives of this line mainly aim to shift the next word probability distribution toward target keywords~\cite{Dathathri2020Plug,pascual-etal-2021-plug-play} or topping up more-target-keywords-contained candidate sequences during generation~\cite{lu-etal-2021-neurologic}. \textcolor{blue}{These decoding methods have made significant progress in balancing the target satisfaction and the text fluency at inference time.}

However, due to the lack of a global and high-level plan for the task, \textcolor{blue}{existing constrained decoding methods usually generate plausible continuations by \textit{greedily} selecting the targets as soon as possible, thereby losing text quality, alignment with natural human expression, and logical control throughout the entire generation process}.
Here, we provide further illustration using the cases in Figure~\ref{fig:intro_case}, where different decoding methods are employed to guide the same PLM for each task (details can be found in the case study sections).
\textcolor{blue}{In CommonGen task (left side), the famous decoding methods {\small \textsc{NeuroLogic}}~\cite{lu-etal-2021-neurologic} and {\small \textsc{Cold}}~\cite{qin2022cold} greedily focus on the targets contributes to quick task satisfaction, however, at the cost of violating a human natural expression:} \textit{we rarely point out the commonsense that we all know such as ``\underline{{\small {\tt wheeled bike}}}''}. In contrast, when we tackle this task, we may follow an implicit \textit{rule} that \textcolor{blue}{not only focuses on the targets but the related concepts that contribute to the construction of a more natural and vivid scenario~\cite{liu2021kg}.}
\textcolor{blue}{Similarly, in the personalized dialogue generation setting (right side of Figure~\ref{fig:intro_case}),
compared to the unconstrained beam search~\cite{lin-etal-2020-commongen}, which only generates user-coherent responses, the constrained decoding method~\cite{lu-etal-2021-neurologic} can meet the task requirements for generating highly personalized replies based on keywords in the persona.
However, frequently copying the keywords in the profile to respond makes the model self-centered, which violates the patterns of everyday conversation. }
A recent study~\cite{xu2022cosplay} also found that even training-based personalized models also suffer from \textit{egocentrism}: they tend to copy the keywords in the target persona by any means to steer the conversation towards their own interests, at the cost of user experience and model interactivity.
In contrast, as personalized individuals responding in daily conversations, we may follow an implicit \textit{rule} of finding common interests: we tend to \textit{selectively} ``copy'' keywords that are relevant to both parties to construct our responses.

\begin{figure*}[!t]
\centering
 \includegraphics[width=\textwidth]{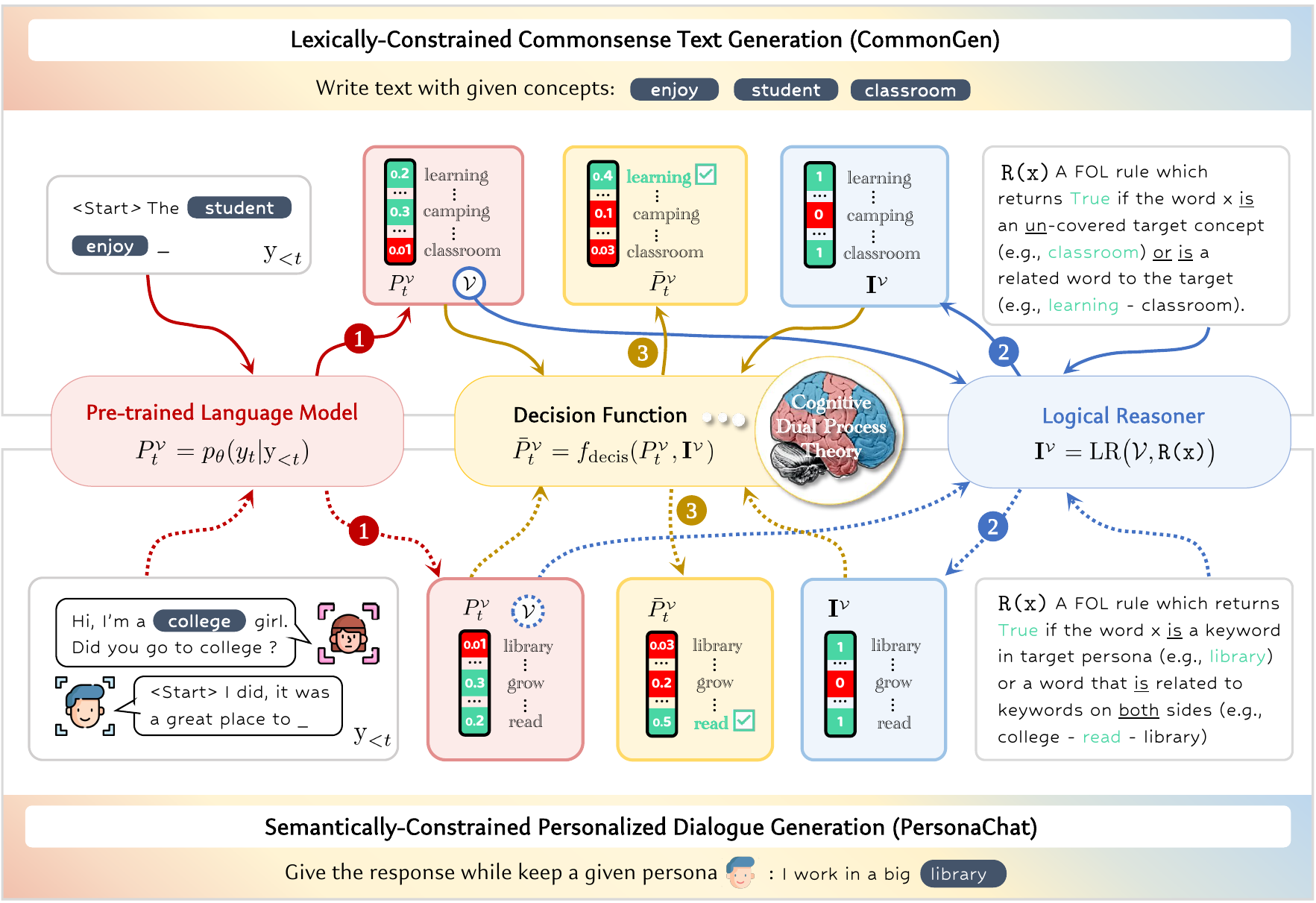}
\caption{
Applying dual-system \modelname{} decoding for CommonGen (top solid arrows) and PersonaChat (bottom dashed arrows) with different rules. 
%
In these two simple examples, {\small {\tt classroom}} (top) and {\small {\tt library}} (bottom) are selected by the logical reasoner because they meet the rule as the target words. However, they are suppressed by the PLM because they violate the intuition of human fluent speaking.
In contrast, {\small {\tt camping}} and {\small {\tt grow}} are good options for the PLM, but bad ones for the logical reasoner. Because their appearance may cause the generation to deviate from the target in the future (e.g., we may not go {\small {\tt camping}} in the {\small {\tt classroom}}). Through the interplay of two systems, \modelname{} will prefer the fluent words that also meet our rules.}
\label{fig:overview}
\vspace{-1em}
\end{figure*}

%
%
%
%
\textcolor{blue}{Therefore, the objective of this paper is to (1) mitigate the greedy focus on target words and (2)
explore a new strategy in constrained decoding methods: guiding pre-trained generative models with logic rules derived from our plans or experiences for the tasks, such as those mentioned in Figure~\ref{fig:intro_case}.
If these can be achieved, the decoded text will not only meet the constrained requirements but also better align with our natural expression and our logic in handling these tasks.}
\textcolor{blue}{To achieve the aforementioned goal, in this work, we first generalize the guiding object from several target words $\texttt{C}$ to a logical rule {\small {\tt R(x)}}. In its simplest form, {\small {\tt R(x)}} can be defined as whether a word {\tt x} is one of $\texttt{C}$, thus reducing into the previous paradigm. However, this distinction allows the new decoding paradigm to employ more sophisticated rules that transforming the encouragement of $\texttt{C}$ into the encouragement of all words {\tt x} that contribute to the task.
Next, to allow the rule signal to flow into the PLM, we take inspiration from dual process theories in cognitive science to explore a neuro-symbolic generative system.} According to the theory, human decision-making is made through the interplay between an intuitive and unconscious System 1 (S1) and a logical and controllable System 2 (S2)~\cite{daniel2017thinking}. Drawing from this, we propose \modelname{}, a rule-controllable decoding framework 
that consists of three parts: 1) a base PLM as S1 to generate fluent text; 2) a First-Order-Logic (FOL) reasoner as S2 to express and evaluate rules; and 3) a \textit{decision function} to merge outputs from two systems during each decoding step.
Figure~\ref{fig:overview} intuitively shows how they collaborate to predict the next for different tasks.

Basically, the rules in our framework can be applied to shift any probability distribution, not only the one for the next prediction in Figure~\ref{fig:overview}, but also the attention distribution over previous words. Modifying the focus of PLM has also been demonstrated to lead to performance improvements~\cite{ji-etal-2022-controlling, dong-etal-2021-fly, Dathathri2020Plug}. 
Therefore, the idea behind \modelname{} can be summarized as follows: although the PLM is a black box, it will produce meaningful and understandable distributions \textcolor{blue}{over words {\small {\tt P(x)}}. We consider these distributions as opportunities to interact with human logic rules {\small {\tt R(x)}} to modify the ``intuitive" behaviors of the neural network.}

\modelname{} is a general rule-controllable framework that can be applied to a variety of tasks. Because during generation, the logical reasoner just calls predicates to calculate the logical vector, delegating how to define them to the users. In experiments, we apply \modelname{} for lexically- and semantically-constrained tasks, leaving the exploration of its other use cases for future work.
For each task above, we first design predicates to describe the rule. Then, \modelname{} is compared with other competitive decoding methods by guiding the same widely-used PLM in the task. Finally, we study the impact of the rule on the final generation.

In summary, our contributions are shown as follows:
\begin{itemize}[leftmargin=5.1mm]
\item We introduce a novel rule-controllable decoding strategy for constrained generation. This strategy allows us to logically program and inject our plan for completing the task, \textcolor{blue}{encouraging not only target words but also all other words that contribute to achieving the task}.
\item We propose \modelname{}, a dual-system decoding framework that integrates a PLM with a logic reasoner for parsing and evaluating FOL rules. \modelname{} also uses a decision function to combine signals from both systems, \textcolor{blue}{allowing logic signals to guide the PLM's focus and predictions at each step, without additional training.}
\item 
\modelname{} framework allows users to program rules according to different tasks. Both quantitative and qualitative results on CommonGen and PersonaChat demonstrate that \modelname{} can generate logically controllable and target-completed text in a more natural and human-like manner.
\end{itemize}

\section{Methodology}
\subsection{Overview}

\modelname{} is a dual-system inspired decoding framework that includes a base PLM for text generation, a logical reasoner for rule interpretation, and a decision function to explicitly model the interplay between logical reasoner and PLM.
%
At each decoding time step $t$, \modelname{} models the next-word distribution $p_t(\ y_t|\text{y}_{<t}, $ {\small {\tt R(x)}}$)$ given both previous words $\text y_{<t}$ and a FOL rule {\small {\tt R(x)}}. As is shown in Figure~\ref{fig:overview}, this process can be viewed as three steps : 
\textbf{\circled{\small1}} given previous words $\text y_{<t}$,  the PLM first produces a probability distribution $P_t^{_\mathcal V}$ over vocabulary $\mathcal{V}$ for the next word prediction (red arrow);
\textbf{\circled{\small2}} then, the logical reasoner applies {\small {\tt R(x)}} to all words in the vocabulary to produce a logical vector $\mathbf I^{_\mathcal{V}}$ that contains truth values (blue arrow). 
\textbf{\circled{\small3}} finally, the decision function $f_{\text{decis}}(P_t^{_\mathcal V},\mathbf I^{_\mathcal{V}})$ combines results from both systems by shifting the distribution toward words with corresponding truth values in $\mathbf I^{_\mathcal{V}}$ which equal to 1 (yellow arrow). The final $\bar P_t^{_\mathcal V}$ will replace the original $P_t^{_\mathcal V}$, steering the direction of generation. 

In the sub-sections below, we first begin with logical reasoner in Section~\ref{sec:s2} and decision function in Section ~\ref{sec:decis}. After demonstrating all the modules, we formulate  the entire decoding process of \modelname{} in Section~\ref{sec:decider}.



\subsection{Logical Reasoner}\label{sec:s2}

\begin{figure*}[!t]
\centering
\includegraphics[width=\textwidth]{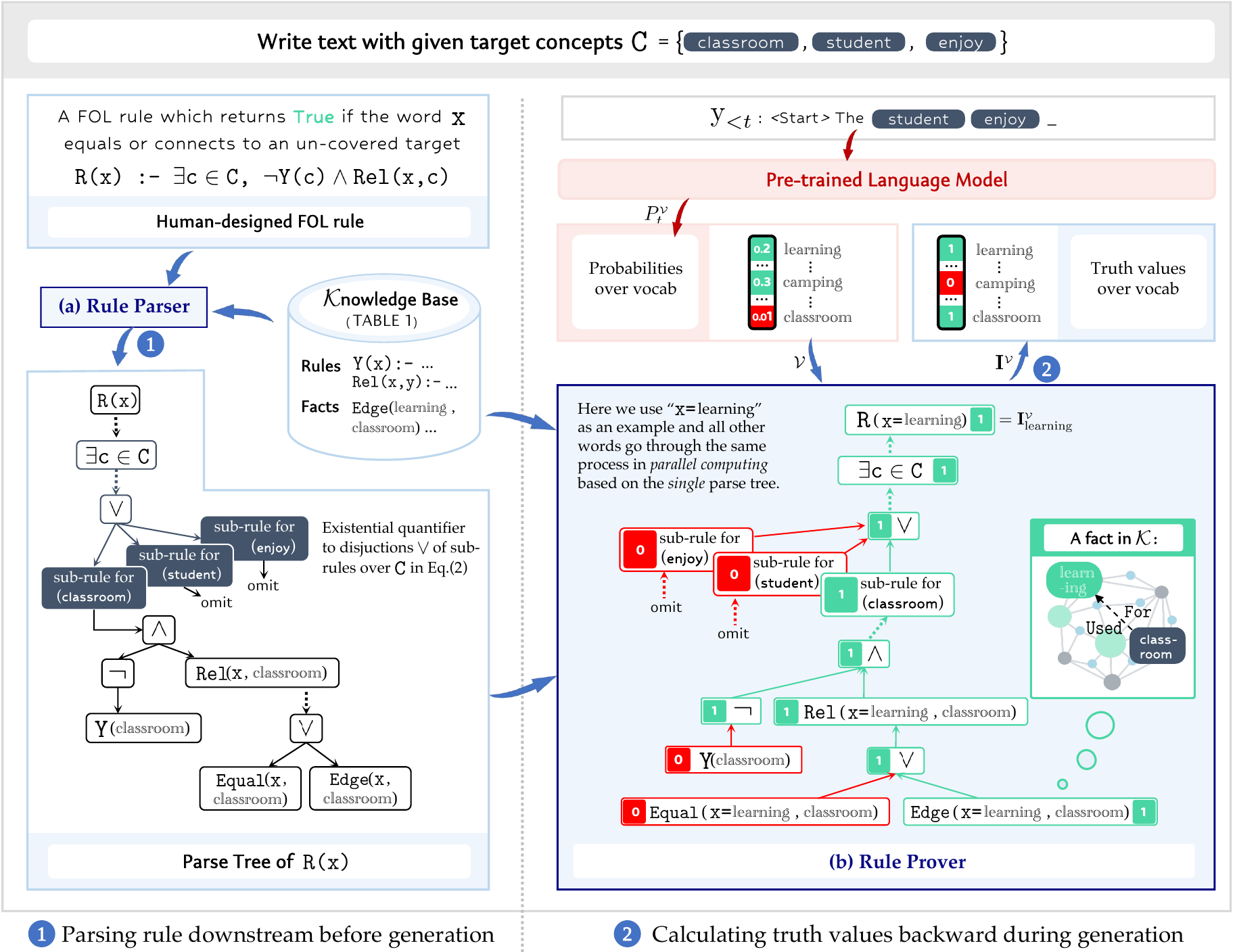}
\caption{
The logical reasoner aims to produce a logic vector over a vocabulary given a user-defined FOL rule. It has a (a) parser to construct a parse tree and a (b) prover to prove whether words meet the rule based on the tree. In this example, though the $\texttt{learning}$ at the lower right corner is not a target ($\texttt{Equal=0}$), it is semantically relevant to a target $\texttt{classroom}$ since there is an edge typed ``used-for'' between them in the knowledge graph ConceptNet ($\texttt{Edge=1}$). Hence, the logic disjunction over them plays the role of selecting not only the targets but also the words that contribute to the targets. These words would have been ignored based on the past alone but incorporate the probability of task completion in the future.}
\label{fig:logical_reasoner}
\vspace{-1em}
\end{figure*}

Figure~\ref{fig:logical_reasoner} gives an overview of logical reasoning with the same example in CommonGen. At each decoding step, the logical reasoner aims to produce a logic vector $\mathbf I^{_\mathcal{V}}$ over a vocabulary. Each element in $\mathbf I^{_\mathcal{V}}$ is a truth value about whether the corresponding word meets the user-defined rule {\small {\tt R(x)}}. The logical reasoner uses a \textit{rule parser} and a \textit{rule prover} for achieving the above process:
\textbf{\circled{\small1}} Before generation, the rule parser first turns the top rule into a parse tree downstream according to the knowledge base $\mathcal{K}$ from the top one all the way to the atomic rules.
\textbf{\circled{\small2}} At each decoding step during generation, the rule prover will traverse the parse tree backward, evaluating the logic vector by grounding the rules to the facts (e.g., $\texttt{x=learning}$). The truth values for all words in the vocabulary are computed in a parallel way.

In this section, we first introduce our knowledge base $\mathcal{K}$ which includes all rules, facts, logic connectives, and quantifiers used in this paper ($\S$~\ref{sec:kb}). Next, we demonstrate how to make an alignment between the word-based predicates and the token-based LM vocabulary ($\S$~\ref{sec:align}). Then, we describe how to turn predicates into their vector-version implementation ($\S$~\ref{sec:vector_functions}). This is necessary for the proving process because the vocabulary is usually very large and all the words in it need to be calculated in parallel. Lastly, we show how the rule parsing and rule proving steps are used to obtain the final logic vector during generation ($\S$~\ref{sec:pp}). 


\begin{table*}[!h]
    \centering
    \caption{The definition and descriptions of rules and facts in our knowledge base $\mathcal{K}$}
    \label{tab:kb}
    \setlength{\tabcolsep}{4pt}
    \renewcommand{\arraystretch}{1.8}
    \begin{tabularx}{\textwidth}{c|lX}
        \hline
        \multicolumn{1}{c|}{\textbf{Type}} & \multicolumn{1}{c}{\textbf{Definition}}                                                                             & \multicolumn{1}{c}{\textbf{Description}}                                           \\
        \hline
        \multirow{4}{*}{\tabcell{Rules for\\ Common-\\Gen}}
            & 1.\ $\texttt{R(x):- } \exists \texttt{c}\in\texttt{C, } \neg \texttt{Y(c)} \wedge \texttt{Rel(x, c)}$ & The word $\texttt{x}$ satisfies the rule if $\texttt{x}$ is related to an un-covered target concept.                                                                                                                                                                                                                                                                                                                                                                                                                                                                 \\
        \cline{2-3}
        & 2.\ $\texttt{Rel(x, y):- } \texttt{Edge(x,y)} \vee \texttt{Equal(x, y)}$                                & Words $\texttt{x}$ and $\texttt{y}$ are related if they are equal or there is an edge between them in a given knowledge graph.                                                                                                                                                                                                                                                                                                                                                                                                                       \\
        \cline{2-3}
        & 3.\ $\texttt{Y(x):- } \exists \texttt{y}\in\texttt{y}_{<t}\texttt{, } \texttt{Equal(x, y)}$     & The word $\texttt{x}$ has been covered in the generated text $\texttt{y}$$_{<t}$ so far if there is a previous word $\texttt{y}$ equals to the word $\texttt{x}$.                                                                                                                                                                                                                                                                                                                                                                                                                                                     \\
        \hline
        \multirow{4}{*}{\tabcell{Rules for\\ Persona-\\Chat}} 
        & 4.\ $\texttt{R(x):- {Persona}(x) $\lor$ {Common}(x)}$                                        & The word $\texttt{x}$ satisfy the rule if $\texttt{x}$ is a persona keyword or the one related to the common interests of both sides.                                                                                                                                                                                                                                                                                                                                                                                                                                 \\
        \cline{2-3}                          
        & 5.\ $\texttt{{Persona}(x):- $\exists$p $\in$ P, {Equal}(x, p)}$                                      & The word $\texttt{x}$ is a persona keyword if there is a keyword $\texttt{p}$ in the given persona description equals to the word $\texttt{x}$.                                                                                                                                                                                                                                                                                                                                                                                                                                                       \\
        \cline{2-3}                          
        & 6.\ $\begin{aligned} 
            \texttt{Common(x):- } 
                 (\ &\exists \texttt{p} \in \texttt{P},\texttt{ Edge(x, p)} )\ \wedge \\
                 (\ &\exists \texttt{u}\in \texttt{U},\texttt{ Edge(x, u)} )
        \end{aligned}$ & The word $\texttt{x}$ is related to the common insterests if $\texttt{x}$ is a bridging word that connects to both sides, i.e., it connects to both a word in the self-persona description $\texttt{P}$ and a word in the user input context $\texttt{U}$.                                                                                                                                                                                                                                                                                                                                                                          \\
        \hline
        \multirow{5}{*}{\tabcell{Facts\\ across\\ Tasks}}&
        7.\ $\Big\{\ \texttt{Equal(w}_\texttt{i}\texttt{, w}_\texttt{j})\ |\  \texttt{w}_\texttt{i}\texttt{w}_\texttt{j}\in s\texttt{, } s\in\mathcal{S}\ \Big\}$                          
        & Words with same stem are seen as equaled, e.g., the word group $s=$ {\scriptsize{\tt \{ran, run, running, runs\}}}. $\mathcal{S}$ denotes word groups with different stems.                                                                                                                                                                                                                                                                                                                                                                                                                              \\
        \cline{2-3}                          & 
        8.\ $\begin{aligned}
        & \Big\{\ \texttt{Edge(v}_\texttt{i}\texttt{,v}_\texttt{j})\ |\  \texttt{Edge(v}_\texttt{i}\texttt{,v}_\texttt{j})\in\mathcal{E}\texttt{, } \texttt{v}_\texttt{i} \texttt{v}_\texttt{j}\in\mathcal{V}\ \Big\}\\
        & \texttt{  or the soft edges:}\\
                                                           &\Big\{\ \texttt{W}(\texttt{v}_\texttt{i}\texttt{,v}_\texttt{j})\ |\  \texttt{W}(\texttt{v}_\texttt{i}\texttt{,v}_\texttt{j})\in\mathcal{W}\texttt{,} \texttt{v}_\texttt{i} \texttt{v}_\texttt{j}\in\mathcal{V}\ \Big\}
                                                       \end{aligned} $ & The facts are from the knowledge graph ConceptNet $\mathcal{G} =(\mathcal{V},\mathcal{E},\mathcal{W})$ where $\mathcal{V}$ denotes its concept verticles, edges $\mathcal{E}$ denote the relevances between concept, and $\mathcal{W}$ are the weights of edges indicating the intensity of the relevances.{\scriptsize{\tt{Edge(v}}}$_\mathrm{i}${\scriptsize{\tt{,v}}}$_\mathrm{j}${\scriptsize{\tt{)}}} are the facts which states that there is relevance between two concepts, such as {\scriptsize{\tt{Edge(rain,umbrella)}}}. A soft verion {\scriptsize{\tt{W(v}}}$_\mathrm{i}${\scriptsize{\tt{,v}}}$_\mathrm{j}${\scriptsize{\tt{)}}} is the normalized weight of {\scriptsize{\tt{Edge(v}}}$_\mathrm{i}${\scriptsize{\tt{,v}}}$_\mathrm{j}${\scriptsize{\tt{)}}} whose truth value ranging from {\scriptsize{\tt{(0, 1)}}} instead of {\scriptsize{\tt{\{0, 1\}}}}. \\
        \hline
    \end{tabularx}
\end{table*}

\subsubsection{First-Order Logic and Rule Knowledge Base}\label{sec:kb}

\vspace{1em}
\noindent\textbf{Rules and Facts.} FOL models the world through 1) objects, such as words in this paper, and 2) relations between objects, such as ``two words are relevant''. Such a relation is represented by \textit{predicate}, a function that maps its arguments to true or false. In this paper, all predicates are defined and stored in the knowledge base $\mathcal K$ in Table~\ref{tab:kb}. The predicates can be further divided into \textit{rules} and \textit{facts}. The rule is denoted with {\small {\tt Head:-Body}} that reads \textit{if} the {\small {\tt Body}} holds \textit{then} the {\small {\tt Head}} holds, where the {\small {\tt Head} is a predicate, the {\small {\tt Body} is multiple sub-predicates with \textit{logical connectives} and \textit{quantifiers}, and {\small {\tt:-}} is logical implication notation.
A fact is a rule whose body always holds and is indicated by {\small {\tt Head}}, which is equivalent to {\small {\tt Head:-true}}. In Table~\ref{tab:kb}, rows 1-6 are rules and 7-8 are facts.

\vspace{1em}\noindent\textbf{Logic Connectives.} 
In order to consider both hard and soft logic~\cite{bach2015hinge} which allows continuous predicates whose output truth values ranging from $[0,1]$ instead of $\{0,1\}$. Therefore, the logic connectives are reformulated as the following equations: 
\begin{equation}\label{eq:connective}
\small
\begin{aligned}
\texttt{P}_1\vee \texttt{P}_2 &= \min(1,\texttt{P}_1+\texttt{P}_2) \\
\texttt{P}_1\wedge\texttt{P}_2 &=(\texttt{P}_1+\texttt{P}_2)\ /\ 2 \\
\texttt{P}_1\  \texttt{\&}\ \texttt{P}_2 &= \max(\texttt{P}_1+\texttt{P}_2-1, 0) \\
\neg\ \texttt{P}_1 & = 1 - \texttt{P}_1 \\
\texttt{P}_1 & \equiv \texttt{P}_2
\end{aligned}
\end{equation}
where $\texttt{P}_i$ is a predicate, $\texttt{\&}$ and $\wedge$ are two different approximations to logical conjunction~\cite{foulds2015latent}. Specifically, $\texttt{\&}$ is a more-hard selection operator (e.g., $\texttt{P}_1 \texttt{\&} \texttt{P}_2=\texttt{P}_2$ when $\texttt{P}_1=1$, and $\texttt{P}_1\texttt{\&}\texttt{P}_2=0$ when $\texttt{P}_1=0$). $\wedge$ is an averaging operator. $\vee$ is the logical disjunction; $\neg$ is the negation operator; $\equiv$ is the logic equivalence indicating that the predicates $\texttt{P}_1$ and $\texttt{P}_2$ have the same truth value.

\vspace{1em}\noindent\textbf{Quantifiers.} 
In certain scenarios, we need quantifiers to formally express the meaning of ``for all'' and ``there is''. For instance, row 1 in Table~\ref{tab:kb} states that if \textit{there is} an un-covered target concept $\texttt{c}$ that is relevant to the word $\texttt{x}$, then the $\texttt{x}$ is the one that meets our expectations.
There are two types of quantifiers: existential quantifier $\exists \texttt{c}\in\texttt{C}, \texttt{P(C)}$ that reads ``there is a $\texttt{c}\in\texttt{C}$ such that $\texttt{P(c)}$'' and universal quantifier $\forall \texttt{c}\in\texttt{C}, \texttt{P(C)}$ that reads ``for every $\texttt{c}$ in $\texttt{C}$ such that $\texttt{P(c)}$''.
To automatically process and calculate quantifiers, we transform existential ($\exists$) and universal ($\forall$) quantifiers equivalently as disjunction ($\vee$) and conjunction ($\wedge$) connectives by iterating the domain $\texttt{C}$, respectively:
\begin{equation}\label{eq:equi}
\small
\begin{aligned}
    \exists \texttt{c}\in \texttt{C}, \texttt{P}(\texttt{c})  \equiv \texttt{P}(\texttt{c}_1) \vee \texttt{P}(\texttt{c}_2) \vee ... \vee \texttt{P}(\texttt{c}_{|\texttt{C}|}) \\
    \forall \texttt{c}\in \texttt{C}, \texttt{P}(\texttt{c})  \equiv \texttt{P}(\texttt{c}_1) \wedge \texttt{P}(\texttt{c}_2) \wedge ... \wedge \texttt{P}(\texttt{c}_{|\texttt{C}|})
\end{aligned}
\end{equation}
where both sides have identical truth-value semantics, $\texttt{P}$ is a predicate, and $\texttt{C}$ is the domain of $\texttt{c}$ (here we use the letter $\texttt{C}$ just to align with the example; it can be any letter). For example in Fig.~\ref{fig:logical_reasoner}, the above equation is used to construct the parse tree.

\vspace{1em}
\subsubsection{Word to Token Alignment}\label{sec:align}
When constructing facts from ConceptNet, we employed two methods to address the potential mismatch between the knowledge graph's word-level triplet nodes and the tokens that the logical reasoner operates on.

\vspace{1em}\noindent\textbf{Preprocessing Alignment.} During the preprocessing of ConceptNet, we primarily focus on relationships between single-word concepts. For multi-word phrases, we decompose them into multiple single-word relationships. For each triplet $(h, r, t)$ which denotes that the head concept $h$ has a relation $r$ with the tail concept $t$, we use the tokenizer to convert both $h$ and $t$ into their corresponding token IDs in the LLM's vocabulary with the parameter {\small{\tt add\_prefix\_space=True}} to maximally prevent words from being broken into sub-units with less semantic information (If either $h$ or $t$ cannot find matching IDs, the triplet is discarded). For the words that are split into multiple tokens, we only consider the first token as the semantic representative of the morphology. 
As a result, each fact is essentially stored as triplets of IDs such as{\small{\tt \ <ID$_{cars}$, ID$_{traffic}$, TruthValue>}}.
    
\vspace{1em}\noindent\textbf{Robust Matching.} We also incorporated the (re)stemming / lemmatization process to extend the facts to ensure that if there is a fact about two words, such as {\small{\tt Edge(car, accident)}}, the other facts involving their different morphological forms are also constructed, such as {\small{\tt Edge(cars, accident)}}, which solves the mismatches due to different morphology during the facts matching phase.

\vspace{1em}
\subsubsection{Vector Version of First-Order Logic}\label{sec:vector_functions}
According to the rules and facts defined in the knowledge base $\mathcal{K}$, traditional tools such as Prolog~\cite{colmerauer1990introduction} can be used to logically ``prove'' whether a word meets a FOL rule through a recursive backward chaining algorithm.
Theoretically, it can be used to produce the logic vector $\mathbf I^{_\mathcal{V}}$ over a vocabulary $\mathcal{V}$ by iterating the above process on every word.
However, in our scenario, the vocabulary for the next prediction is very large, e.g., the vocabulary of GPTs can be up to tens of thousands of words. Therefore, this motivates us to calculate  logic over the entire vocabulary in parallel instead of word by word.

In order to achieve this, we extend the predicate $\texttt{P(x)}$ to its vector version implementation $\texttt{P}\texttt{(}\mathbf{X}\texttt{)}$, where its variable $\mathbf{X}$ is a bag-of-word vector over the vocabulary instead of a single word $\texttt{x}$. Moreover, $\texttt{P}\texttt{(}\mathbf{X}\texttt{)}$ returns a logic vector of all truth values for $\mathcal{V}$ instead of a single truth value for $\texttt{x}$. This logic vector can be viewed as a new bag of words that is fed into other predicates as input.
Therefore, the job of the logic reasoner can be viewed as using vector-version predicates to filter the bag of all words $[1]^{\mathcal{V}}$ into the bag of words that meet our rules, i.e., the final logic vector $\mathbf I^{_\mathcal{V}}$.
In terms of the implementation of predicates, we turn all for loops into vector and matrix calculations. Here we use the implementation of $\texttt{Edge(x, p)}$ at row 6 in Tab.~\ref{tab:kb} as an example. Its vector implementation is defined as 
\begin{equation}
    \texttt{Edge(}\mathbf{X}\texttt{,p)} = \mathbf{X}_{|\mathcal{V}|\times1}\times \mathbf{E}_{|\mathcal{V}|\times|\mathcal{V}|}[:, \texttt{\small{index}\_of(p)}]
\end{equation}
where $\mathbf{E}$ is an \textit{adjacency matrix} with a dimension of $|\mathcal V|\times|\mathcal V|$ (in practice it is supported by Pytorch sparsity matrix). Each element in the matrix represents whether there is an edge between the row and column words. The matrix is pre-processed ($\S$~\ref{sec:impl}) from ConceptNet~\cite{speer2017conceptnet}. In the above equation, the element-multiplication ($\times$) between $\mathbf{X}$ and one column of $\mathbf{E}$ indexed by $\texttt{p}$ aims to obtain a logic vector of whether the bag of words in $\mathbf X$ connects to the keyword $p$ in the persona profile. Then $\exists \texttt{p}\in\texttt{P, } \texttt{Edge(}\mathbf{X}\texttt{, p)}$ can be transformed into the logic disjunctions (Eq.~\ref{eq:equi}):
\begin{equation}
\texttt{Edge}(\mathbf{X}, \texttt{p}_1) \vee \texttt{Edge}(\mathbf{X}, \texttt{p}_2) \vee ... \vee \texttt{Edge}(\mathbf{X}, \texttt{p}_{|\texttt{P}|})
\end{equation}
where all connectives, such as $\vee$, are implemented by Pytorch operators over vectors according to the Equation~\ref{eq:connective}.

\vspace{1em}
\subsubsection{Rule Parser and Rule Prover}\label{sec:pp}
The logical reasoner aims to produce a logic vector of truth values about whether the words in a vocabulary satisfy a user-defined FOL rule. As is shown in Figure~\ref{fig:logical_reasoner}, such a process can be further divided into two steps.

\vspace{1em}
\noindent\textbf{Parsing Step.} 
Before generation, the rule parser first turns the rule into a parse tree ($\mathcal{T}$).
Given the knowledge base, the tree is constructed from the top rule $\texttt{R(x)}$ as the root to the atomic rules as the leaves, such as $\texttt{Equal}$ and $\texttt{Edge}$. Our FOL parser is built on a open-source syntax parser framework. The quantifier node in the tree is then changed into conjunctions or disjunctions of sub-trees based on Eq.~\ref{eq:equi}. 
In addition, each node of the tree is linked to a function that is used to evaluate the truth value of this node (described in Sec.~\ref{sec:vector_functions}). These functions are called in the proving step.

\vspace{1em}
\noindent\textbf{Proving Step.} 
At each time step during generation, the rule prover will traverse the parse tree $\mathcal{T}$ backward, computing the truth values for the entire vocabulary $\mathcal{V}$ at each node. The truth values of all leaf nodes are evaluated by calling their predicate functions, and the truth values of non-leaf nodes are computed by logical connectives (solid arrows) or kept by logical equivalents (dashed arrows), as shown in Fig.~\ref{fig:logical_reasoner}. In this example, the proving step only shows the calculation on a single word, such as $\texttt{Edge(x=learning, classroom)}$. However, the prover actually does it in parallel by calculating the leaf $\texttt{Edge(}\mathbf{X}\texttt{, classroom)}$ whose $\mathbf{X}$ is initialized by $[1]_{|\mathcal{V}|\times 1}$, a bag of all words over $\mathcal{V}$. This vector-version predicate will return a logic vector over the entire vocabulary (described in Sec.~\ref{sec:vector_functions}). The above rule-proving process is concluded in Algorithm~\ref{algo:traversal}. 
Lastly, the logic vector $\mathbf I^{_\mathcal{V}}$ obtained from the root node $\mathcal{T}$ is the final output of the logical reasoner.

\subsection{Decision Function}\label{sec:decis}
From the above section, 
for any distribution ${P^{\mathcal{V}}}$ over a vocabulary ${\mathcal{V}}$, the logical reasoner can produce a logical vector $\mathbf{I}^{\mathcal{V}}$ over the same ${\mathcal{V}}$ about whether each word in it meets a user-defined rule:
\begin{equation}
\mathbf{I}^{\mathcal{V}} = \texttt{LR}(\mathcal{V}, \texttt{R(x)})
\end{equation}
Then the decision function $f_{\texttt{decis}}$ uses a logic vector to shift a probability distribution towards words whose truth value equals one:
\begin{equation}\begin{aligned}
      &\bar P^{\mathcal{V}}= \ f_{\texttt{decis}}(P^{\mathcal{V}}, \mathcal{\mathbf{I}^{\mathcal{V}}}; \alpha)\\
       &\ = \texttt{softmax}\Bigl(\texttt{invsoft}(P^{\mathcal{V}})+\mathbf{I}^{\mathcal{V}}\times (\alpha P^{\mathcal{V}})\Bigr) 
\end{aligned}\end{equation}
where $\texttt{invsoft}(P^{\mathcal{V}})$ (inverse softmax) stores the pre-activated scores, and $\alpha$ is a positive parameter for the intensity of rule control.
The intuition is that we want to 1) increase the probabilities of words whose corresponding truth values equal one by boosting their pre-activated scores before softmax, and 2) balance the rule control and text quality: the positive original probability $P^{\mathcal{V}}$ automatically decides the magnitude of the boost, which avoids making a big adjustment to an originally small probability even if the word meets the rule.
Therefore, only the words that are not bad choices $P^{\mathcal{V}}$ and also meet the rule can gain a good boost from this function.
From this perspective, the decision function plays the role of combining both decisions from neural and logic systems. The resultant $\bar P^{\mathcal{V}}$ is the perturbed distribution used to substitute the original one, adjusting the``intuitive'' behaviours of the PLM.

\makeatletter
\newcommand{\algorithmfootnote}[2][\footnotesize]{%
  \let\old@algocf@finish\@algocf@finish
  \def\@algocf@finish{\old@algocf@finish
    \leavevmode\rlap{\begin{minipage}{\linewidth}
    #1#2
    \end{minipage}}%
  }%
}
\makeatother
\begin{algorithm}
    \footnotesize
    \LinesNumbered
    \SetAlgoNlRelativeSize{-1}
    \caption{Recursive Proving}
    \algorithmfootnote{\ $^*\ $The $\triangle$ represents the logical operator such as $\vee$, and $\wedge$.}
    \label{algo:traversal}
    \KwData{the parse tree $\mathcal{T}$, a vocab $\mathcal V$ , the knowledge base $\mathcal{K}$}
    \KwResult{logic vector $\mathbf{I}^{\mathcal{V}}$ }
    
    \SetKwFunction{ProvingTraversal}{ProvingTraversal}
    \SetKwFunction{Evaluate}{evaluate}
    
    \SetKwProg{Fn}{Function}{:}{}
    \Fn{\ProvingTraversal{root, vocab, knowledge base}}{
        $\mathbf{I}^{\mathcal{V}} \leftarrow$ \Evaluate{root, vocab, knowledge base}\;
        \KwRet $\mathbf{I}^{\mathcal{V}}$\;
    }
    
    \SetKwProg{Fn}{Function}{:}{}
    \Fn{\Evaluate{node, vocab, knowledge base}}{
        \If{node is the leaf node}{
            truth\_values $\leftarrow$ Evaluating the truth values over $\mathcal{V}$ by calling rules, facts and logical operators which are defined in $\mathcal{K}$ \;
            \KwRet truth\_values\;
        }
        \Else{
            truth\_values\_list $\leftarrow$ empty list\;
            \For{each child in children of node}{
                truth\_values $\leftarrow$ \Evaluate{child, $\mathcal{V}$, $\mathcal{K}$}\;
                truth\_values\_list.append(truth\_values)\;
            }
            truth\_values $\leftarrow$ $\triangle$$^*$(truth\_values\_list)\;
            \KwRet truth\_values over $\mathcal V$\;
        }
    }
\end{algorithm} 
\subsection{\modelname{} Decoding}\label{sec:decider}
The idea behind \modelname{} is that during generation, the PLM will produce some probability distributions, and we just see these distributions as the interface for communicating with human logic signals. In this section, we show how logical reasoner perturbs these distributions and formulate the entire decoding process.
For the base PLM, we adopt the currently most commonly used structure: the pre-trained transformer decoder~\cite{vaswani2017attention} with $L$ transformer blocks (e.g., the GPT family~\cite{liu2023summary,radford2018improving, radford2019language, zhang-etal-2020-dialogpt}). At each time step $t$, the generation process of PLM can be viewed as understanding previous words $\textbf{y}_{<t}$ and target words $C$ in the attention distribution, then predicting the next word $y_t$ in the prediction distribution. We formulate them one by one through forward propagation along with the figure~\ref{fig:decis-decoding}.

\vspace{1em}
\noindent\textbf{Shifting Attention Distribution over Previous Words.}
At each time step $t$, we denote all the previous words to the current time as $\textbf{y}_{<t}$. For a self-attention layer on certain head in transformer, we represent the key-value pairs of previous layer for the input $\mathbf{y}_{<t}$ as $\mathbf{K}_{<t}\mathbf{V}_{<t}$, then the self-attention distribution $A_{<t}$ is calculated by the product of the current query $q_{t-1}$ with all previous keys $K_{<t}$ (the yellow part in Figure~\ref{fig:decis-decoding}). 
The $A^{<t}$ will be shifted to $\Bar{A}^{<t}$ when the logical vector $\mathbf{I}^{<t}$ is provided by the logical reasoner given a FOL rule $\texttt{R(x)}$:
\begin{equation}\label{eq:self_att}
\begin{aligned}
A^{<t} =& \texttt{softmax}\left(\frac{\ q_{t-1} \cdot (\mathbf{K}_{<t})^{\top}}{\sqrt{d}}\right) \\
\mathbf{I}^{<t} =&\ \mathrm{LR}(\mathbf{y}_{<t}, \texttt{R(x)})\\
\bar{A}^{<t}=& \ f_{\texttt{decis}}(A^{<t}, \mathbf{I}^{<t})\\
\end{aligned}\end{equation}
The perturbed \textcolor{purple}{$\bar{A}^{<t}$} will replace the original ones to re-assign the attentions more on the previous words that satisfy our rule.

\vspace{1em}
\noindent\textbf{Shifting Attention Distribution over Target Words.}\label{sec:p2}
An additional operation to the standard self-attention layer is that we also individually feed the target words $C$ into the PLM, to pre-produce stack of positional-invariant key-value pairs~\cite{carlsson2022fine} $\textbf{K}_C\textbf{V}_C = \mathrm{PLM}(C)$ (the red part in Figure~\ref{fig:decis-decoding}). Then the attention distribution $A^{C}$ over target words is calculated by the product of the current query $q_{t-1}$ with $\textbf{K}_{C}$. 
Similarly, $A^{C}$ will be shifted to $\Bar{A}^{C}$ when a logical vector is provided by the  $\mathrm{LR}$:
\begin{equation}\label{eq:concept_att}
\begin{aligned}
A^{C} =& \texttt{softmax}\left(\frac{\ q_{t-1} \cdot (\textbf{K}_C)^{\top}}{\sqrt{d}}\right) \\
\mathbf{I}^{C} =&\ \mathrm{LR}(C, \texttt{R(x)})\\
\Bar {A}^C=& \ f_{\texttt{decis}}(A^C, \mathbf{I}^{C})\\
\end{aligned}\end{equation}
The perturbed $\Bar {A}^C$ will replace the original ones to introduce more information from the target words that satisfy our rule.
Next, to combine the attentions of previous and target words, the final attention distribution of this layer is the concatenation of them: $A_{[C:<t]} = [\Bar{A}^C : \Bar{A}^{<t}]$.
With the dot production to their values $\textbf{V}_{[C:<t]}=[\textbf{V}_{C} : \textbf{V}_{<t}]$, the hidden state is calculated by $h_t = f(A_{[C:<t]}\cdot \textbf{V}_{[C:<t]})$ ($f$ contains the immediate transformer operations~\cite{vaswani2017attention} such as residual connections, layer normalization and feed-forward network for the concise expression). Finally, $h_t$ is used to produce the query, key and value for the next self-attention layer.

\vspace{1em}
\noindent\textbf{Shifting Prediction Distribution for Next Word.}\label{sec:p3}
The probability distribution $P_{t}$ for the next word is generated from the last layer hidden state $h^{L}_{t}$. 
This distribution over LM vocabulary $\mathcal V$ will be shifted to $\bar{P}^{\mathcal{V}}$ (blue part in Figure~\ref{fig:decis-decoding}) when a logical vector is given:
\begin{equation}\label{eq:pred}
\begin{aligned}
P_t^{\mathcal{V}} =& \texttt{softmax}(W\cdot h^L_t)\\ 
\mathbf{I}^{\mathcal{V}} =&\ \mathrm{LR}(\mathcal{V}, \texttt{R(x)})\\
\bar{P}_t^{\mathcal{V}}=& \ f_{\texttt{decis}}(P_t^{\mathcal{V}}, \mathbf{I}^{\mathcal{V}})\\
\end{aligned}\end{equation}
%
Finally, to balance the proportion of rule-relevant and target words, \modelname{} employs beam search with pruning \cite{post-vilar-2018-fast} to top up candidates that include both target words~\cite{lu-etal-2021-neurologic} and rule-satisfying words.




\subsection{Complexity Analysis}
\textcolor{blue}{The logic reasoner evaluates each word in the vocabulary according to a recursive proving algorithm (described in Sec.~\ref{sec:pp}, proving step).Given that the number of defined rules and sub-rules per rule are both very small, the time complexity for evaluating each word is approximately \( O(1) \). Furthermore, we adopt parallel computation, allowing the logic reasoner to evaluate all words simultaneously, so evaluating the entire vocabulary incurs no additional time complexity.Then the Decision Function adjusts the probability distribution for the next word(described in Sec.~\ref{sec:decis}).Due to the parallel computation, the time complexity for adjusting the whole vocabulary is also \( O(1) \).In summary, \modelname{} introduces logical computation to shift the probability distribution of traditional beam search. The improvement uses parallel computing which does not incur any additional time overhead. Therefore, for generating a sequence of N words, the overall time complexity of \modelname{} is \( O(Nk) \), which is equivalent to that of beam search. }

\section{Experiments I: Constrained Commonsense Generation on CommonGen}\label{task:common}
In the constrained commonsense generation task, the model is expected to generate a sentence that describes a common scenario using a given set of target concepts $C=\{c_1, ..., c_n\}$. However, the challenge lies in the fact that in order to build a common scenario, the model may need to reason out relevant concepts in addition to the given ones~\cite{liu2021kg}. 
Therefore, we first design a rule $\texttt{R(x)}$ that is shown in the row 1 in Tab.~\ref{tab:kb} to control the decision-making process of \modelname{}: \textit{the model should focus on not only the un-covered targets but also the related ones that induce them.} At each decoding step, \modelname{} injects rule signals into PLM at both attention and prediction distributions, steering the generative direction.

\begin{figure}[!t]
\centering
\includegraphics[width=\columnwidth]{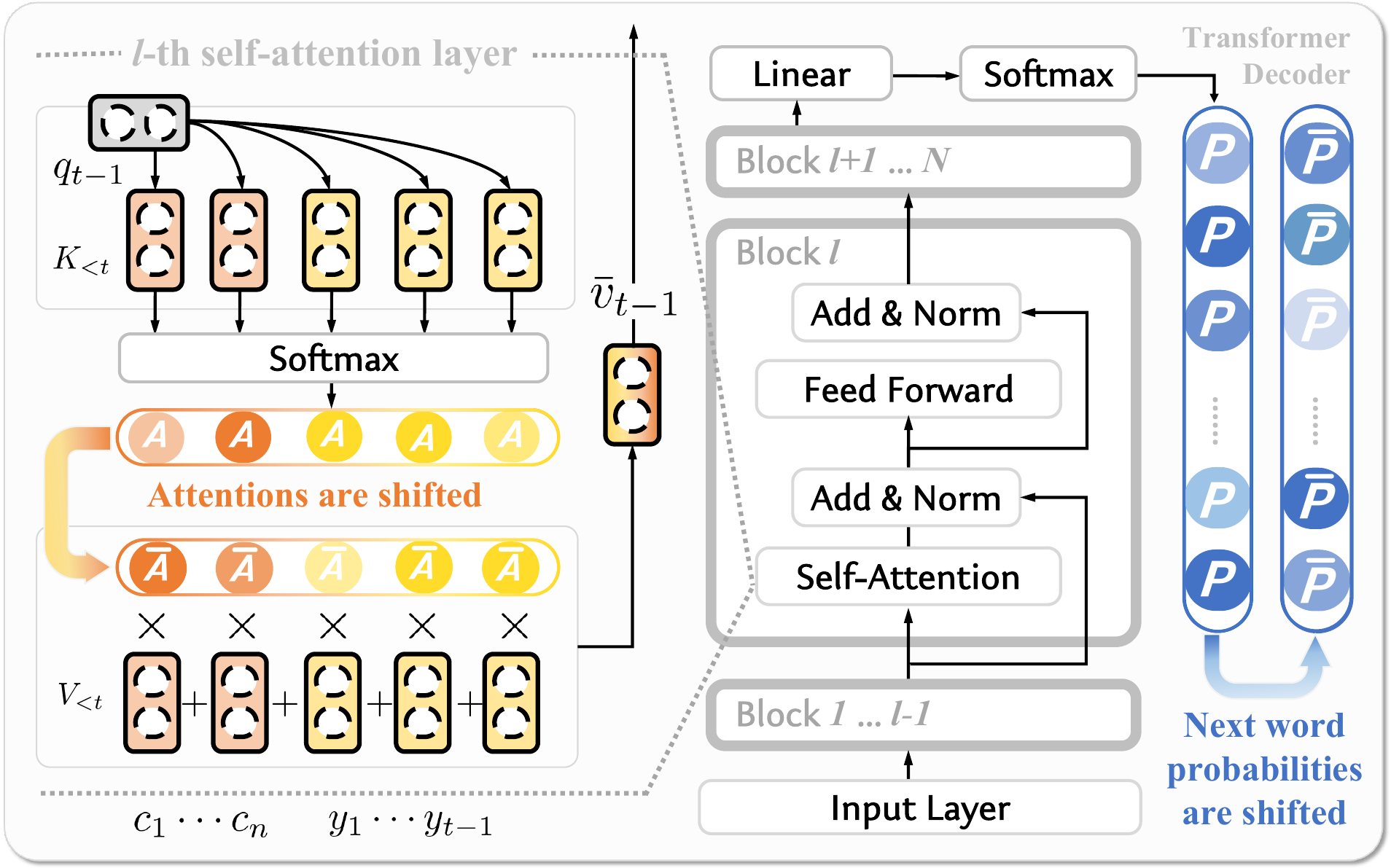}
\caption{Illustration of \modelname{} decoding applied to the base transformer-structured language model during generation.}
\label{fig:decis-decoding}

    \vspace{-1em}
\end{figure}

\begin{table*}[!t]
\setlength{\tabcolsep}{2.5pt}
	\centering
 	\caption{Performance of the different unconstrained and constrained decoding methods for the CommonGen benchmark}
	\resizebox{\textwidth}{!}{
	\begin{tabular}{l c c c c c c c}
	\toprule
		\multicolumn{1}{l}{\multirow{-1}{*}{\bf{Decoding Method}}}  & 
		\multicolumn{1}{l}{\multirow{-1}{*}{\bf{Constrained}}}  & 
		\multicolumn{1}{c}{\multirow{-1}{*}{\bf \ \ BLEU(\%) \ \ }} &
		\multicolumn{1}{c}{\multirow{-1}{*}{\bf \ \ ROUGE(\%) \ \ }} &
		\multicolumn{1}{c}{\multirow{-1}{*}{\bf METEOR(\%)}} &
		\multicolumn{1}{c}{\multirow{-1}{*}{\bf \ \ CIDEr\ \ }} &
		\multicolumn{1}{c}{\multirow{-1}{*}{\bf \ SPICE(\%)\ }} &
		\multicolumn{1}{c}{\multirow{-1}{*}{\bf Coverage(\%)}}\\
	\midrule
            Top-k Sampling~\cite{fan2018hierarchical}  &\XSolidBrush&  18.73 &1.62 &10.25 &0.54 &7.65 &8.3 \cr
		  Beam Search Decoding~\cite{lin-etal-2020-commongen}& \multirow{-1}{*}{\XSolidBrush} & 19.56 &1.54 &11.22 &0.87 &7.21 &8.2 \cr
\cdashline{1-8}
            Keyword2Text Decoding~\cite{pascual-etal-2021-plug-play} &\CheckmarkBold&16.37 &33.34 &23.23 &10.46 &21.46 &84.5 \cr 
            PPLM Decoding~\cite{Dathathri2020Plug} &\CheckmarkBold&4.23 &21.35 &15.94 &6.97 &14.33 &19.1\cr 
            Constrained Beam Search \cite{anderson-etal-2017-guided} &\CheckmarkBold& 20.47 & 37.83 & 28.15 & 12.64 & 27.17 &96.5 \cr
            Grid Beam Search \cite{hokamp-liu-2017-lexically} &\CheckmarkBold& 17.86 & 37.04 & 26.53 & 11.44 & 26.16 &96.3\cr
            Dynamic Beam Allocation ~\cite{post-vilar-2018-fast} &\CheckmarkBold&18.26 &37.15 & 27.25 & 12.26 & 26.27 &96.4 \cr
            TSMH Decoding ~\cite{zhang2020language} &\CheckmarkBold&10.72 &24.70 & 14.50 & 3.60 & 15.40 & 71.5 \cr
            COLD Decoding ~\cite{qin2022cold} &\CheckmarkBold& 23.76 & 33.56 & 17.21 & 13.08 & 24.35 & 94.5\cr
             AttnM Decoding ~\cite{dong2021fly} &\CheckmarkBold& 26.80 & 39.32 & 28.02 & 13.71 & 23.94 & 81.9\cr
            \textsc{NeuroLogic} Decoding ~\cite{lu-etal-2021-neurologic} &\CheckmarkBold&{24.53} &{41.84} &{29.57} &{14.24}&{27.53} & {96.7}\cr
		\modelname{} Decoding (Ours) & \CheckmarkBold & \bf{27.03}  &\bf{43.14} &\bf{30.13} &\bf{15.24}&\bf{28.52} &\bf{97.1}\cr
		\bottomrule
	\end{tabular}}
	\label{tab:auto_common}
\end{table*}

\subsection{Experiment Setup}
\noindent\textbf{Datasets.} 
We validate our method on the CommonGen test set. It is designed for generative commonsense reasoning and lexically-constrained tasks, which requires models to compose given discrete concepts (e.g., {dog, frisbee, catch, throw}) into coherent sentences depicting daily scenarios. The dataset contains 35,141 concept sets (32,651 in train, 993 in validation, and 1,497 in test) associated with 77,449 sentences. The average size of the concept sets in the test set is 4.04, and the average sentence length is 13.34 words.

\vspace{1em}
\noindent\textbf{Evaluation Metrics.} Following CommonGen~\cite{lin-etal-2020-commongen}, we adopt widely-used metrics~\cite{tian-etal-2024-sheng} for text generation, which focus on (1) n-gram overlap such as BLEU~\cite{papineni2002bleu}, ROUGE~\cite{lin2004rouge}, and METEOR~\cite{banerjee2005meteor}, (2) concept-oriented metrics such as CIDEr~\cite{Vedantam_2015_CVPR} and SPICE~\cite{anderson2016spice}, and (3) concept coverage rate~\cite{lin-etal-2020-commongen}. For human evaluation, we sample 100 test examples with concepts and generation pairs. Three crowdsource occupational annotators are asked to give a score $\in\{1,2,3\}$ for quality, commonsense, and informative. Specifically, for quality, the sentence may be: (1) neither well-formed nor fluent, (2) understandable but awkward, or (3) human-level quality; For common sense, the sentence may be (1) counter, (2) not counter, or (3) very consistent with our common sense; For informative, the sentence may 1) only focus on the given concepts, (2) be more vivid, but introduced information is not necessary for building a better scene, such as adjectives, or (3) build a more informative and commonsensible scenario contributed by the extra information.

\vspace{1em}
\noindent\textbf{Base Model and Baselines.}
We compare our method with the classical unconstrained and competitively constrained decoding strategies based on the same open-source PLM, the GPT-2-large, for text generation:
Top-k sampling~\cite{fan2018hierarchical} and beam search decoding~\cite{lin-etal-2020-commongen}  is the most commonly used unconstrained decoding.
For the constrained ones, PPLM~\cite{Dathathri2020Plug} changes the neural hidden states (key-value pairs) with the gradients from an external bag-of-words model to increase the probability of the target words.
Keyword2Text~\cite{pascual-etal-2021-plug-play} directly shifts the probability distribution over the vocabulary towards the words with similar word embeddings to the target ones.
Constrained Beam Search~\cite{anderson-etal-2017-guided} is an approximate search algorithm to enforce the target concepts over the resulting output sequences with a finite-state machine;
Grid Beam Search~\cite{hokamp-liu-2017-lexically} (GBS) extends the beam search to incorporate the grouping method that groups together hypotheses by the number of constraints satisfied.
Dynamic Beam Allocation~\cite{post-vilar-2018-fast} incorporates the pruning method into the GBS.
\textcolor{blue}{COLD~\cite{qin2022cold} unifies constrained generation as specifying constraints through an energy function, then performing efficient differentiable reasoning over the constraints through gradient-based sampling. AttnM~\cite{dong2021fly} dynamically redistribute sentence-level attention weights by injecting task-specific priors in Transformer blocks for different downstream tasks. TSMH~\cite{zhang2020language} integrates a tree search algorithm into the proposal process of MCMC to explore candidates that satisfy more constraints.}\textsc{NeuroLogic}~\cite{lu-etal-2021-neurologic}, as the master of this type of method, incorporates both grouping and pruning to select the hypotheses that satisfy more target words.

To further investigate the effectiveness of \modelname{} across models of different capability levels, we extend our experiments beyond GPT-2-large~\cite{lin-etal-2020-commongen} to advanced open-source large language models, including LLaMA2-13B~\cite{touvron2023llama} and LLaMA3.1-8B~\cite{dubey2024llama}. We instruct these models to perform the CommonGen task with the prompt template \textit{``Use the given words to make a short sentence that is consistent with commonsense. Words: \{...\}''.} 
This comparative study allows us to verify whether the performance improvements of \modelname{} persist across models with varying capability levels.

\begin{table}[!t]
\setlength{\abovecaptionskip}{4pt}
	\centering
 	\caption{Human evaluation on the CommonGen dataset}
	\resizebox{\columnwidth}{!}{
	\begin{tabular}{l c c c c}
	\toprule
		
		\multicolumn{1}{l}{\multirow{-1}{*}{\bf{Decoding Method}}}  & 
		\multicolumn{1}{c}{\multirow{-1}{*}{\bf{Quality}}}  & 
		\multicolumn{1}{c}{\multirow{-1}{*}{\bf Common.}} &
		\multicolumn{1}{c}{\multirow{-1}{*}{\bf Infor.}} &
		\multicolumn{1}{c}{\multirow{-1}{*}{\bf Average}}\\
	\midrule
            Keyword2Text~\cite{pascual-etal-2021-plug-play}        &1.36 &1.48 &{1.92} &1.59  \cr
            \textsc{NeuroLogic}~\cite{lu-etal-2021-neurologic}  &{2.16} &{2.30} &1.68 &{2.05}  \cr
            \modelname{} (Ours)   &\bf2.21 &\bf2.33 &\bf2.21 &\bf2.25  \cr    
		\bottomrule
	\end{tabular}}
	\label{tab:human_common}
\end{table}

\subsection{Results and Analysis.}
\noindent\textbf{Automatic and Human Evaluation.}
The automatic and human evaluation results are shown in Table \ref{tab:auto_common}, Table \ref{tab:human_common} and Table \ref{tab:additional_exp}. We can see: (1) \modelname{} outperforms all other previously constrained decoding methods with respect to all metrics, making a good trade-off between the text quality and task completion (Coverage).
\textcolor{blue}{(2)In Table \ref{tab:auto_common}, \modelname{} performs well on the Coverage metric, indicating its ability to cover all target words to complete the task. \modelname{} also performs well on metrics that measure the similarity between the generation and human reference, such as BLEU, ROUGE, and METEOR. This demonstrates that \modelname{} can produce text that closely resembles human-generated patterns. More importantly, \modelname{} also achieves high scores on more concept-oriented matching metrics such as CIDEr and SPICE. This indicates that the generated text can depict more natural and realistic scenarios, as different scenarios in a sentence are usually related to the keywords or concepts they contain.
(3) The human evaluation results indicate that \modelname{} can use the target-relevant information introduced by the rule to build more naturally expressed and vivid text (Info.) without decreasing the quality of sentences (Quality) or violating commonsense (Common.). 
(4) Compared with \neurologic{}~\cite{lu-etal-2021-neurologic}, the improvement of Common. is limited. The reason is that while the \neurologic{}~\cite{lu-etal-2021-neurologic} model tends to generate some obvious texts such as ``wheeled bike'', it still adheres to commonsense because bikes do have wheels in reality. However, such expressions tend to reduce Info, making it less effective in generating more natural and vivid scenarios compared to \modelname{}. (5) In Table \ref{tab:additional_exp}, the effectiveness of \modelname{} demonstrates a negative correlation with the model's inherent capabilities: it achieves more significant improvements on baseline models like GPT-2-large~\cite{lin-etal-2020-commongen}, while showing relatively smaller gains on more capable models like Llama3.1-8B~\cite{dubey2024llama}. This phenomenon can be attributed to the vast difference in pre-training data:   Llama2-13B~\cite{touvron2023llama} and Llama3.1-8B~\cite{dubey2024llama}. were trained on substantially larger datasets that likely included CommonGen-like data, where models learn to compose coherent text from given concepts. Consequently, these models may have already acquired similar capabilities during pre-training, limiting the potential for further improvement through \modelname{}. This finding highlights \modelname{}'s practical value for domain-specific applications where deploying large models is impractical. For instance, by improving the capabilities of lightweight models, \modelname{} offers a viable solution when facing computational constraints, making it particularly valuable in scenarios with limited resources or strict deployment requirements.}

\begin{table}[!t]
\setlength{\tabcolsep}{2.8pt}
	\centering
 	\caption{Performance comparison of \modelname{} applied to different pre-trained lightweight or large language models. All metrics except CIDEr are presented on a percentage scale.}
	\resizebox{\columnwidth}{!}{
	\begin{tabular}{l c c c c c c c}
	\toprule
		\multicolumn{1}{l}{\multirow{-1}{*}{\bf{Model}}}  & 
		\multicolumn{1}{c}{\multirow{-1}{*}{\bf \textsc{Bleu}}} &
		\multicolumn{1}{c}{\multirow{-1}{*}{\bf \textsc{Rouge}}} &
		\multicolumn{1}{c}{\multirow{-1}{*}{\bf \textsc{Mete}.}} &
		\multicolumn{1}{c}{\multirow{-1}{*}{\bf \textsc{Cide}r}} &
		\multicolumn{1}{c}{\multirow{-1}{*}{\bf \textsc{Spice}}} &
		\multicolumn{1}{c}{\multirow{-1}{*}{\bf \textsc{Cov}.}}\\
        \midrule
            GPT-2-large~\cite{lin-etal-2020-commongen}& 19.56 &1.54 &11.22 &0.87 &7.21 &8.2 \cr
            $\hookrightarrow$ + \modelname{} & \bf{27.03}  & \bf{43.14} & \bf{30.13} &\bf{15.24}& \bf{28.52} & \bf{97.1}\cr
	\midrule
              LLaMA2-13B~\cite{touvron2023llama} & {32.28} & {44.13} & {28.86} & {14.98} & {26.43} & {93.84} \cr
              $\hookrightarrow$ + \modelname{} & \bf{33.04} & \bf{44.72} & \bf{29.43} & \bf{15.49} & \bf{27.02} & \bf{97.56} \cr
	\midrule
            Llama3.1-8B~\cite{dubey2024llama} & {38.07} & {51.72} & {31.45} & {15.37} & {30.24} & {96.14} \cr
            $\hookrightarrow$ + \modelname{}& \bf{38.56} & \bf{52.28} & \bf{31.75} & \bf{15.46} & \bf{30.31} & \bf{98.41}\cr
	\midrule
            GPT-3.5~\cite{brown2020language}& 37.44 &49.01 &29.05 &15.13 &27.74 &95.16 \cr
            GPT-4~\cite{openai2023gpt}& 42.87 &52.98 &32.56 &16.49 &30.95 &96.67 \cr
		\bottomrule
	\end{tabular}}
    \label{tab:additional_exp}
    
\end{table}

\vspace{1em}
\noindent\textbf{Variant and Ablation Study.}
As shown in Table \ref{tab:ablation_common}, we conduct a variant study for two versions of the \modelname{} and an ablation study to see the influence of  shifting different distributions by the rule on the performance. Some key observations are: 1) The influence of \texttt{soft} predicate logic is important. The soft version allows the logical reasoner to $\textit{weighted}$ control the semantics of generation by treating word relevance differently. 
2) Cutting off the perturbation on the prediction distribution results in the most deteriorated performance, since $\Bar{P^{\mathcal{V}}_t}$ degrades into ${P^{\mathcal{V}}_t}$ (Eq.~\ref{eq:pred}), it will lose the power to directly select the next words that satisfy our rules.
 3) The decreases in performance for $A^C$ (Eq.~\ref{eq:concept_att}) and $A^{<t}$ (Eq.~\ref{eq:self_att}) also suggests that properly adjusting the attentions over target words and previously generated words also contribute to the final results.

\vspace{1em}
\noindent\textbf{Case Study.}
Figure~\ref{fig:case_common} shows two cases compared with the \textsc{NeuroLogic} as baseline and we observe that: baseline tends to generate boring sentences mainly constrained on the target words while \modelname{} can generate more commonsense and informative sentences, benefiting the rules to focus on externally related concepts. In addition, the semantic and senario  behind depicted by the decider is closer to the human described ones such as a person is hit by the ball casually in the Case 2.

\section{Experiments II: Personalized Response Generation on PersonaChat}\label{task:persona}

In the personalized response generation task, the model is expected to generate a response $\mathbf{y}$ based on a user-input utterance $\mathbf{u}$. In addition, $\mathbf{y}$ is required to be semantically consistent with a given target persona $S=\{\mathbf{s}_1, ..., \mathbf{s}_n\}$ which consists a set of sentences to describe the interests (e.g., ``I like football'') or occupations (e.g., ``I am a teacher''). Intuitively, ``copying'' the persona keywords or concepts into response appears to produce plausible personalized responses.
However, focusing too much on self persona makes the model more egocentric~\cite{xu2022cosplay}, which may violate our daily conversational scenario: When talking with a partner, we usually follow a persona selection strategy: finding the common ground to chat.
Therefore, we develop a rule as is shown at the row 1 in Tab.~\ref{tab:kb}: \textit{the model should focus on not only the self persona keywords but also the words falling to the common interests of both sides, even if they do not directly appear in the target persona.}
During generation, the \modelname{} will inject the rule signals into PLM at both attention and prediction distributions according to Sec.~\ref{sec:decider}, influencing the generation.

\begin{table}[!t]
\setlength{\abovecaptionskip}{4pt}
\setlength{\tabcolsep}{2.8pt}
 	\caption{Variant and ablation study on CommonGen. Since the edge predicate in Tab.~\ref{tab:kb} has both a soft version $\texttt{W(v}_\texttt{i}\texttt{,v}_\texttt{j})$ by default and a hard version $\texttt{Edge(v}_\texttt{i}\texttt{,v}_\texttt{j})$, which results in the final \modelname{} having two variants. $\hookrightarrow$ - means we ablate the influence of shifting different distributions by the rule one by one, which is achieved by setting the logical vector $\mathbf{I}$ as zeros to cut off rule signals. Hence the distributions will degrade into their original ones in turn.}
	\centering
	\resizebox{\columnwidth}{!}{
	\begin{tabular}{l c c c c}
	\toprule
		
		\multicolumn{1}{c}{\bf{Variant}}  & 
		\multicolumn{1}{c}{\multirow{-1}{*}{\bf ROUGE(\%)}} &
		\multicolumn{1}{c}{\multirow{-1}{*}{\bf METEOR(\%)}} &
		\multicolumn{1}{c}{\multirow{-1}{*}{\bf SPICE(\%)}} &
		\multicolumn{1}{c}{\multirow{-1}{*}{\bf Cover.(\%)}}\\
	\midrule
            \modelname{} ($\texttt{Hard}$)    &38.33 &27.91 &26.32 & 88.1\cr\midrule
            \modelname{} ($\texttt{Soft}$)   &\bf43.14 &\bf30.13 &\bf28.52& \bf97.1 \cr
            $\hookrightarrow$ - $\Bar A^{<t}$ in Eq.~\ref{eq:self_att}
            &40.21 &28.42  &26.25 &94.5\cr
            $\hookrightarrow$ - $\Bar A^{C}$ in Eq.~\ref{eq:concept_att}
            &38.11 &27.45 &24.12 & 89.2\cr
            $\hookrightarrow$ - $\Bar P_t^{\mathcal{V}}$ in Eq.~\ref{eq:pred}
            &29.33 &21.23  &19.07 & 69.1\cr     
		\bottomrule
	\end{tabular}}
	\label{tab:ablation_common}
    
\end{table}

\subsection{Experiment Setup.}
\noindent\textbf{Datasets.}  
We conduct experiments on PersonaChat~\cite{zhang2018personalizing}, a dialogue dataset designed for semantically-constrained tasks. Each sample consists of a dialogue history with no more than 15 utterances and a persona description containing 4-6 profile sentences. The dataset contains 8,939/1,000 dialogues conditioned on 1,155/100 personas for the train/dev set. We report all results on the latter. Note that we also present the experimental results on the revised dataset where \textit{the original personas are rephrased, generalized, or specialized}~\cite{zhang2018personalizing} because there is a danger that during the dataset construction, humans will unwittingly copy persona information either verbatim or with significant word overlap. Therefore, such a revised PersonaChat makes the task more challenging because it requires the generalization ability of methods to reason out more relevant information instead of just copying the content in the persona descriptions.

\vspace{1em}
\noindent\textbf{Evaluation Metrics.}
The semantically constrained task is different from CommonGen: the gold responses not only literally copy the words in the persona but also construct the response from persona-relevant contents. Therefore, we should consider both the lexicon overlap and the semantic similarity in the metrics. Specifically, to evaluate the quality of the generated text, we measure the difference between the decoded response and the gold reference by the cumulated 2-gram BLEU~\cite{papineni2002bleu}, 4-gram NIST~\cite{cao2022model} (weighted BLEU), ROUGE-L~\cite{lin2004rouge}, F1 score~\cite{xu2022cosplay} (harmonic mean of precision and recall), and BERT-score~\cite{bert-score} (semantic version of F1 computed by the DeBERTa V2 large model~\cite{he2021deberta}). 
For the evaluation of controllability, we designed a persona consistency score which is termed as Persona C$^2$, which leverages a referee model to predict semantic consistency between the generation \textbf{y} and persona sentences of both sides $p_i$:
\begin{equation}\text{C}^2(\textbf{y}) = \sum_{p}\text{NLI}(\textbf{y},p), p\in \{S, U\}
\end{equation}
\begin{equation}
\label{formula:11}
\begin{aligned}
    \text{NLI}(\textbf{y},p_i)&=\left\{
    \begin{aligned}
    -1 ,\text{if} & \ \textbf{y}\ \text{contradicts}\ p_i, \\
    0  ,\text{if} & \ \textbf{y}\ \text{is irrelevant to}\ p_i, \\
    1  ,\text{if} & \ \textbf{y}\ \text{entails}\ p_i.
    \end{aligned}
    \right.\\
\end{aligned}
\end{equation}
where the NLI is a pre-trained RoBERTa model~\cite{liu2019roberta} fine-tuned with the dialogue inference dataset DNLI. The NLI model achieves a test set accuracy of 88.2\% reported by~\cite{welleck-etal-2019-dialogue}. The $\textbf{y}$ is a generated utterance, and the $p_i$ is one sentence in the persona description.
Different from the traditional persona C score~\cite{song-etal-2021-bob, madotto-etal-2019-personalizing}, C$^{2}$ score also considers the consistency of the partner's persona sentences $U$ which is also loaded from the dataset. Therefore, having a higher C$^{2}$ score means that the generated responses are controlled by the common interests of both parties.

For human evaluation, we sample 150 test examples with persona, user context, and generation triplets. Three crowdsourced occupational annotators are asked to give a score $\in\{1,2,3\}$ for quality, consistency, and coherence.
The evaluation protocol for each aspect is shown as follows: Consis. measures whether the response is consistent with the target persona. The response may be 1) contradictory, 2) irrelevant, or 3) consistent with the given persona descriptions. Cohere. measures whether the response is coherent with the user's context. The response may be 1) incoherent, 2) utterance-level coherent, or 3) both utterance-level and persona-level coherent, which means the response not only follows what the user just said but also focuses on the user's persona information shown in the context and finds the common interests to give a reply.

\vspace{1em}
\noindent\textbf{Base Model and Baselines.} 
\textcolor{blue}{For comparisons with \modelname{}, we selected the widely-used unconstrained decoding methods, beam search, as well as the representative constrained decoding methods PPLM~\cite{Dathathri2020Plug}, AttnM~\cite{dong2021fly}, and  \textsc{NeuroLogic}~\cite{lu-etal-2021-neurologic} from Experiment I. All methods are equipped with two open-source base models: 1) Dialogpt-large~\cite{zhang-etal-2020-dialogpt}, a large-scale pre-trained response generation model trained on the 147M Reddit corpus without fine-tuning on the target task. 2) \textsc{P}$^2$\textsc{BOT}~\cite{liu2020you}, a transmitter-receiver model fine-tuned on the PersonaChat dataset to explicitly model the understanding between interlocutors.}

\begin{table*}[!t]
\setlength{\abovecaptionskip}{4pt}
	\centering
 	\caption{Performance of different (un)constrained decoding methods on both \textbb{O}riginal and \textbb{R}evised PersonaChat datasets}
	\resizebox{\textwidth}{!}{
    \begin{tabular}{l c c c c c c c c c}
	\toprule
		
		\multicolumn{1}{c}{\multirow{-1}{*}{\bf{Decode Method}}}  & 
		\multicolumn{1}{c}{\multirow{-1}{*}{\bf Data}} &
        \multicolumn{1}{c}{\multirow{-1}{*}{\bf Base}} &
  	\multicolumn{1}{l}{\multirow{-1}{*}{\bf{Constr.}}}  &
   
		\multicolumn{1}{c}{\multirow{-1}{*}{\bf BLEU(\%)}} &
		\multicolumn{1}{c}{\multirow{-1}{*}{\bf NIST}} &
		\multicolumn{1}{c}{\multirow{-1}{*}{\bf ROUGE(\%) }} &
		\multicolumn{1}{c}{\multirow{-1}{*}{\bf F1(\%)}} &
		\multicolumn{1}{c}{\multirow{-1}{*}{\bf{BERT(\%)}}} &
		\multicolumn{-2}{c}{\multirow{-1}{*}{\bf Persona} C$^2$}\\
\midrule
    Beam Search Decoding~\cite{lin-etal-2020-commongen} & &&\XSolidBrush&{15.78}& 0.71 &11.64& 7.89& 47.22& 0.32\cr
    %
    PPLM Decoding~\cite{Dathathri2020Plug} &&&\CheckmarkBold&{11.25} & {0.76} &{11.75}& {7.25}& {47.25}&{0.51}\cr
    AttnM Decoding ~\cite{dong2021fly}&&&\CheckmarkBold&{15.31}& {0.70} &{12.08}& {7.81}& {50.01}&{0.48}\cr
    
    \textsc{NeuroLogic} Decoding~\cite{lu-etal-2021-neurologic} &&&\CheckmarkBold&15.63& {0.83} &{12.33}& {9.62}& {50.12}&{0.54}\cr
    \modelname{} Decoding (Ours) & {\multirow{-5}{*}{\textbb{O}}}& {\multirow{-5}{*}{Dialogpt~\cite{zhang-etal-2020-dialogpt}}} 
    &\CheckmarkBold&\bf16.59& \bf{0.91} &\bf{13.42} &\bf{10.77}&\bf{50.83}&\bf{0.66}\cr
\cdashline{1-10}
    Beam Search Decoding~\cite{lin-etal-2020-commongen} &&&\XSolidBrush&14.23& 0.79 &10.83& 8.71 & 47.08& 0.36 \cr
    %
    PPLM Decoding~\cite{Dathathri2020Plug} &&&\CheckmarkBold&{10.92} & {0.71} &{11.32}& {7.95}& {46.82}&{0.49}\cr
    AttnM Decoding ~\cite{dong2021fly}&&&\CheckmarkBold&{14.08 }& {0.67} &{11.40}& {8.65}& {49.07}&{0.41}\cr
    
    \textsc{NeuroLogic} Decoding~\cite{lu-etal-2021-neurologic} &&&\CheckmarkBold&
    {14.64}& {0.74} &{11.81}& {8.98}& {48.24} & {0.48} \cr
    \modelname{} Decoding (Ours) & {\multirow{-5}{*}{\textbb{R}}}& {\multirow{-5}{*}{Dialogpt~\cite{zhang-etal-2020-dialogpt}}} &\CheckmarkBold& \bf15.43& \bf{0.81}&\bf{12.92}&\bf{9.81}& \bf{50.11}& \bf{0.62}  \cr
\cdashline{1-10}
    %
    Beam Search Decoding~\cite{lin-etal-2020-commongen} &&&\XSolidBrush&{16.35}&{0.94}&{12.72}&{14.55}&{49.51}&{0.53}\cr
    %
    PPLM Decoding~\cite{Dathathri2020Plug} &&&\CheckmarkBold&{11.74} & {0.98} &{12.79}& {14.36}& {49.53}&{0.62}\cr
    AttnM Decoding ~\cite{dong2021fly}&&&\CheckmarkBold&{15.88}&{0.93} &{12.98}& {14.48}& {51.73}&{0.60}\cr
    \textsc{NeuroLogic} Decoding~\cite{lu-etal-2021-neurologic} &&&\CheckmarkBold&
    {16.23}&{1.01}&{13.32}&{16.11}&{51.78}&{0.67} \cr
    %
    \modelname{} Decoding (Ours) & {\multirow{-5}{*}{\textbb{O}}}& {\multirow{-5}{*}{\pbot{}~\cite{liu2020you}}} &\CheckmarkBold& \bf{17.01}&\bf{1.11}&\bf{14.15}& \bf{17.02}& \bf{52.32}&\bf{0.76}  \cr\bottomrule
\end{tabular}}
\label{tab:auto_persona}
\end{table*}

\begin{table}[!t]
\setlength{\abovecaptionskip}{4pt}
\setlength{\tabcolsep}{2.8pt}
	\centering
  	\caption{Variant and ablation study on PersonaChat. Since the edge predicate in Tab.~\ref{tab:kb} has both a soft version $\texttt{W(v}_\texttt{i}\texttt{,v}_\texttt{j})$ by default and a hard version $\texttt{Edge(v}_\texttt{i}\texttt{,v}_\texttt{j})$, which results in the final \modelname{} having two variants. $\hookrightarrow$ - means we ablate the influence of shifting different distributions by the rule one by one, which is achieved by setting the logical vector $\mathbf{I}$ as zeros to cut off rule signals. Hence the distributions will degrade into their original ones in turn.}
	\resizebox{\columnwidth}{!}{
	\begin{tabular}{l c r c c}
	\toprule
		
		\multicolumn{1}{c}{\bf{Variant}}  & 
		\multicolumn{1}{c}{\multirow{-1}{*}{\bf BLEU(\%)}} &
		\multicolumn{1}{c}{\multirow{-1}{*}{\ \ \bf F1(\%) \ }} &
		\multicolumn{1}{c}{\multirow{-1}{*}{\bf \textsc{BERT(\%)}}} &
		\multicolumn{1}{c}{\multirow{-1}{*}{\bf Persona C$^2$}}\\
	\midrule
            \modelname{}($\texttt{Hard}$)   &15.83&10.22 &50.63& 0.52 \cr\midrule
            \modelname{}($\texttt{Soft}$)   &\bf16.59 &\bf10.77 &\bf50.83& \bf0.66 \cr
            $\hookrightarrow$ - $\Bar A^{<t}$ in Eq.~\ref{eq:self_att}
            &16.11 &10.42  &50.23 &0.59\cr
            $\hookrightarrow$ - $\Bar A^{C}$ in Eq.~\ref{eq:concept_att}
            &15.52 &9.74 &49.54 & 0.53\cr
            $\hookrightarrow$ - $\Bar P_t^{\mathcal{V}}$ in Eq.~\ref{eq:pred}
            &14.93 &9.18  &48.13 & 0.48\cr
    \bottomrule
	\end{tabular}}
	\label{tab:ablation_persona}
\end{table}

\begin{table}[!t]
 \caption{Human Evaluation on the PersonaChat dataset }
\setlength{\abovecaptionskip}{4pt}
\setlength{\tabcolsep}{7.5pt}
	\centering
	\resizebox{\columnwidth}{!}{
	\begin{tabular}{l c c c c}
	\toprule
		
		\multicolumn{1}{l}{\multirow{-1}{*}{\bf{Method}}}  & 
		\multicolumn{1}{c}{\multirow{-1}{*}{\bf{Quality}}}  & 
		\multicolumn{1}{c}{\multirow{-1}{*}{\bf Consis.}} &
		\multicolumn{1}{c}{\multirow{-1}{*}{\bf Cohere.}} &
		\multicolumn{1}{c}{\multirow{-1}{*}{\bf Average}}\\
	\midrule
            Beam Search~\cite{lin-etal-2020-commongen} &2.22 & 2.15 & \underline{2.15} & 2.17\cr
            \textsc{NeuroLogic}~\cite{lu-etal-2021-neurologic}  &\underline{2.23} & \bf2.45 & 1.95 & \underline{2.21}\cr
            \modelname{} (ours)   &\bf2.33 & \underline{2.43} & \bf2.28 & \bf2.35\cr    
		\bottomrule
	\end{tabular}}
	\label{tab:human_persona}
    
    \vspace{-1em}
\end{table}

\begin{figure}[!t]
\centering
\includegraphics[width=\columnwidth]{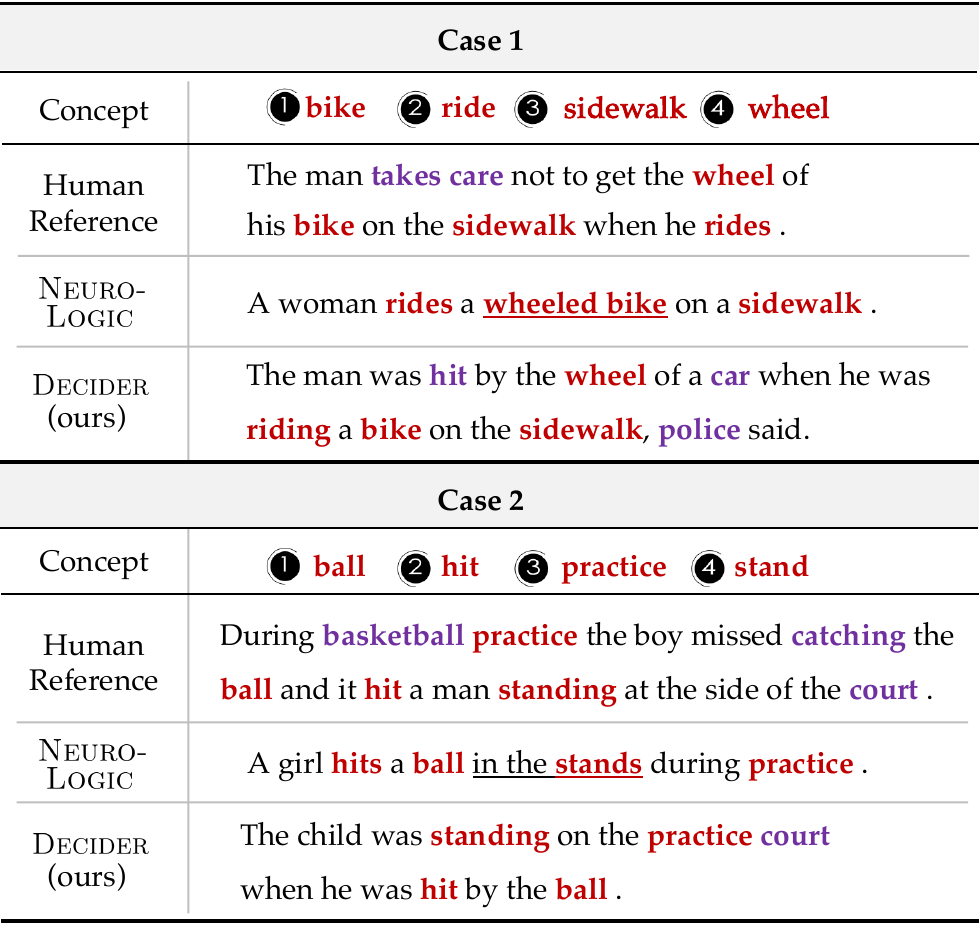}
\caption{
Case study for the CommonGen. Underlined words violate an important commonsense that we rarely tell commonsense or details out unless they are special for the current scene. Benefiting from the rule at row 1 in the Tab.~\ref{tab:kb}, \modelname{} has the ability to guess this special scene at human-level with the external information in purple.}
\label{fig:case_common}

    \vspace{-1em}
\end{figure}

\subsection{Results and Analysis.}

\noindent\textbf{Automatic and Human Evaluation}
The automatic and human evaluation results are shown in Table~\ref{tab:auto_persona} and Table~\ref{tab:human_persona}. The key observations are: (1) \modelname{} outperforms both canonical and constrained decoding for all metrics, making a good trade-off between persona consistency (C$^2\uparrow$) and response quality (other metrics). (2) A good performance on C$^2$ indicates that \modelname{} can follow our rule to find the common interests to chat, because only when consistent with the persona across both parties can the response achieve a high score. In addition, such a change in response pattern influenced by the rule makes the response closer to our real scenario, which explains the improvements in quality metrics that measure the difference between the generation and the gold references.
(3) \modelname{} can also perform well on semantic-oriented similarity metrics such as BERT and C$^{2}$ scores due to the relevant concepts recalled by the high-level rule $\texttt{R(x)}$. Compared to the baseline that focused on the exact target words such as $\textsc{NeuroLogic}$, this advantage would be more pronounced in the revised dataset, where the persona is revised to test the generalization of methods.\textcolor{blue}{(4) \modelname{} can efficiently guide both fine-tuned (\pbot{}~\cite{liu2020you}) and non-fine-tuned (Dialogpt~\cite{zhang-etal-2020-dialogpt}) base models, indicating its generalization over base models.
(5) Even if the base model (\pbot{}~\cite{liu2020you}) has learned the generation pattern through the training data of the task, \modelname{} can still further enhance the performance of the generated text by explicitly embedding rules.
(6) Unlike other baselines, \modelname{} completes the task (\csquare $\uparrow$) without compromising the quality of the text, as all other scores surpass the unconstrained baseline (Beam Search).} (7) The human evaluation results indicate that \modelname{} can follow our rule to build more conversational scenarios by both expressing our interests and taking care of the partner (Cohere.$\uparrow$ and Consis.$\uparrow$) without decreasing the generation quality.

\vspace{1em}
\noindent\textbf{Variant and Ablation Study}
We conduct the variant and ablation studies in Table~\ref{tab:ablation_persona}. Some key observations are: (1) When compared to the hard variant, the soft predicate calculus still plays an important role in performance, indicating that treating relevances between words differently has a positive influence on controlling the more accurate semantics of responses.
(2) Similar to the trend on CommonGen, all the perturbations in the three probability distributions by the rule contribute to the final performance. This indicates that shifting attention or prediction distribution can also change the generation's direction towards the way that is consistent with our rule.

\label{sec:experiments}

\vspace{1em}
\noindent\textbf{Case Study}
Figure~\ref{fig:case_persona} gives two cases and we observe that (1) The responses of Beam Search and \textsc{Neurologic} are either over-focus on the partner or self persona while \modelname{} can do a good balance. (2) \modelname{} can select the common interest (e.g., read and reading in case 1) or explore new concepts in the common ground (e.g., \underline{library}$\rightarrow$\{read, book, school\}$\leftarrow$ \underline{college} in case 2). This is mainly due to rule 6 in Table~\ref{tab:kb}, which makes a concept more competitive if it connects both parties.
\modelname{} can choose the appropriate concepts to construct responses with the strategy: finding the common interests. Blue and purple words are persona and external concepts respectively. The common interests are underlined.

\begin{figure}[!t]
\centering
\includegraphics[width=\columnwidth]{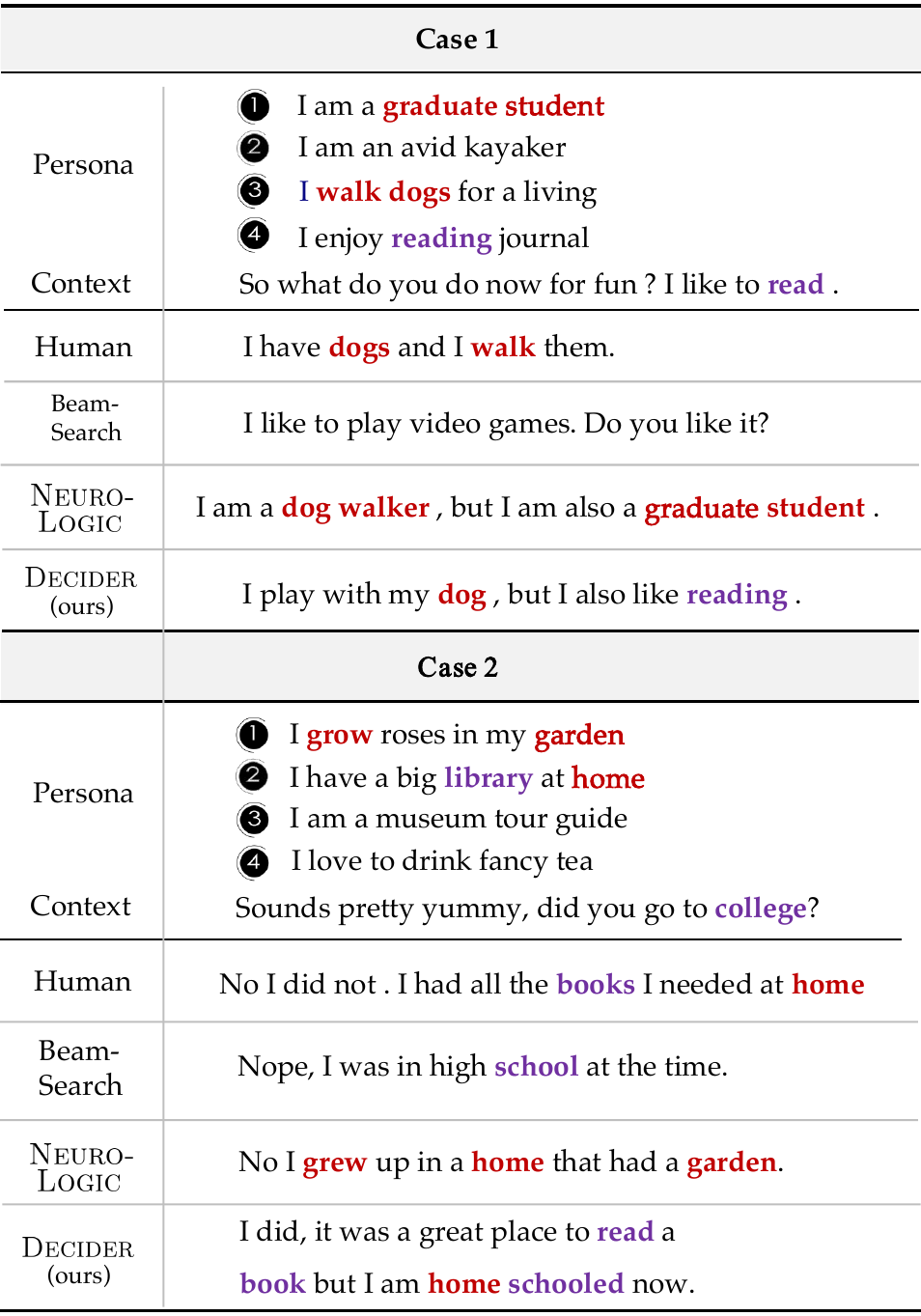}
\caption{
In PersonaChat, two scenarios between the interlocutors that have one common interest are selected for illustration. Following the rule at row 4 in the Tab.~\ref{tab:kb}, \modelname{} has the ability to recognize the common interests and control the responses relevant to the intersection (purple) of the self-persona and the partner-persona in the user input context.}
\label{fig:case_persona}
\end{figure}

\section{Related Work}
\noindent\textbf{Neuro-Symbolic Methods in Natural Language Generation.} Neuro-Symbolic AI aims to integrate the symbolic knowledge into neural generation methods to complement the strengths of each field~\cite{susskind2021neuro}. The symbolic knowledge can be classified into two types: the knowledge graph (KG) such as ConceptNet and the logic rules, such as First-Order Logic.
For KG-enhanced generation. \cite{xu2022cosplay} uses the concept relations from KG to improve the personalized dialogue generation.
\cite{li2020knowledge} inject concepts with higher emotion intensity values from KG into the model for empathetic dialogue generation. 
\cite{shen2024cbt} retrieve domain-specific knowledge and integrates it into large language models to construct a virtual cognitive behavioral therapy models for mental health counseling.
\cite{ji2020language} employs graph attention embedding to encode sub-graphs in pre-trained models for dialogue reasoning. 
Recently, for the other hand, many attempts have been made to inject logic rules in NLP tasks in the way of knowledge regularization~\cite{zhou2021clinical} or knowledge distillation~\cite{hu-etal-2016-harnessing}.
However, this line of research for the generation field remains unexplored. \cite{nye2021improving} and ~\cite{shu-etal-2021-logic} first introduce the semantic parser as the symbolic system to improve consistency and coherence of the generated text. \textsc{NeuroLogic}~\cite{lu-etal-2021-neurologic} is the first to introduce the First-Order Rule to controllable text generation. 
All of the above works are different from ours in that they introduce \textit{either} knowledge graphs or logic rules into the neural generation, while our decoding method introduces \textit{both}  of them: we view the \textit{concrete edges} in KG as the facts of the \textit{general rule} $\texttt{R(x)}$. These facts play the role of evidence for $\texttt{R(x)}$ to query or do reasoning on them.

\vspace{1em}
\noindent\textbf{Constrained Decoding Strategies}
Constrained decoding strategies for directly guiding PLMs without training have sparked a surge in research.
\textsc{NeuroLogic}~\cite{lu-etal-2021-neurologic} guides generation based on whether the candidates contain certain \textit{concrete} target words or not while we guide generation based on whether the candidates satisfy a \textit{high-level} rule (with the variable $\texttt{x}$ to enable implication). Another difference is that our rule introduces the reasoning over the external knowledge graph. 
PPLM methods~\cite{Dathathri2020Plug, xu2021change} changes the neural hidden states with the gradients from an external attribute model while we change the hidden states by combining logical signals in decision function. Keyword2Text~\cite{pascual-etal-2021-plug-play} shifts the probability distribution with similar word embeddings to the target, while we use the general rules to reason out the relevant words that contribute to approach the target.
\textcolor{blue}{A significant difference between the above constrained approaches and this paper is that, instead of guiding the model using indicators of whether specific words are mentioned, \modelname{} introduces the variable \(x\) to represent words in a general form, using whether a word \(x\) satisfies a certain predicate \(\texttt{R(x)}\) to guide the generation, thereby encouraging all words that contribute to or satisfy the human-customized rules rather than only encouraging the specific words.}

\vspace{1em}
\noindent\textbf{Steering Generation by Attention Perturbation.} 
Some decoding methods aim to directly perturb attention on the fly in either 1) a post-activated manner, such as PPLM~\cite{Dathathri2020Plug} which control the generation attributes by using the gradients to shift the original attention, or 2) a pre-activated manner, such as work from~\cite{dong-etal-2021-fly} that perturbed pre-activated attention scores to alleviate text degeneration. Our work borrows the idea that different attentions can lead to different generations and explores a new post-and pre-activated way: our decision function uses the original distribution and a logical vector to perturb the pre-activated scores of the new one.

\subsection{Implementation Details}\label{sec:impl}
The System 1 is built on GPT-2-Large~\cite{radford2019language}, LLaMA2-13B~\cite{touvron2023llama} and LLaMA3.1-8B~\cite{dubey2024llama} for CommonGen and Dialogpt~\cite{zhang-etal-2020-dialogpt} for PersonaChat of HuggingFace transformers~\cite{wolf2019huggingface} and run on one single NIVDIA V100 GPU. All predicate calculus in the System 2 are based on the matrix computations implemented by Pytorch~\cite{paszke2019pytorch}. 
For the hyper-parameters, the batch size is 10; the beam size is 20 for CommonGen and 10 for PersonaChat, which are same for all other baselines that are built on the beam search. All key parameters of \{$\alpha_1$, $\alpha_2$, $\alpha_3$, $\rho$, $k$\} are set as \{12, $24$, $24$, $0.6$, $16$\} for CommonGen and \{12, $24$, $48$, $0.4$, $8$\} for PersonaChat. 
To preprocess facts from ConceptNet, for each triplet $(h, r, t)$ which denotes that the head concept $h$ has a relation $r$ with the tail concept $t$, we use the pre-process pipeline from~\cite{ji2020language} (tokenization, lemmatization, stop-word and black-word filtering) to process both $h$ and $t$ to get $h'$ and $t'$. Then we set $\mathbf{E(h', t')=E(t', h')} = 1$ for $\texttt{Hard}$ or $=w(r)$ for $\texttt{soft}$ predicates. In soft version which the \modelname{} uses by default, $w(r)\in(0,1)$ is the rescaled weight for the relation $r$ in graph.

\section{Limitations}
    During the LLM's generation, certain pieces $\texttt{x}$ (either already generated or to be generated) may satisfy certain predicate rules $\texttt{R(x)}$. When these rules are met, the \modelname{} will enhance the attention or prediction of these pieces at inference. 
    However, the current fine-grained version of \modelname{} only considers single words as $\texttt{x}$ without considering longer linguistic units, leaving more sophisticated improvements to accommodate phrase- and sentence-level predicates for future work.

\section{Conclusions}
In this paper, we propose \modelname{}, a novel decoding strategy for constrained language generation that can be controlled by any FOL rules as we desire.
\modelname{} is driven by the dual systems with a decision function to let the rule signal flow into the PLM at each decoding step.
Extensive experimental results on widely-used benchmarks demonstrate that our proposed \modelname{} significantly outperforms competitive baselines by using well-designed rules to guide the PLM's generative direction towards the targets.

\section{Acknowledgements}
We would like to express sincere gratitude to all the reviewers and meta-reviewers for their insightful and constructive feedback. Besides, we would also
like to thank many researchers for their valuable comments which
greatly improved the quality of our paper: Deng Cai, Leyang Cui, Xinting Huang, Xianyang Wang, Haiyun Jiang, Jieshuo Zhang, Shuming Shi, Xiuling Liu. Furthermore, this
project was funded in part by the Postdoctoral Fellowship Program of CPSF (No. GZC20233403), the National Natural Science Foundation of China (No. 
62450100, 
62227807,
62376030,
62272044,
62373057,
62402041,
62476127,
62406163), 
the China Postdoctoral Science Foundation (No. 2024M764142, 
2021M700423, 2024T171131), 
and the Beijing Natural Science Foundation~(No. L243034).

\bibliographystyle{IEEEtran}
\bibliography{decider.bib}

\begin{thebibliography}{10}
\providecommand{\url}[1]{#1}
\csname url@samestyle\endcsname
\providecommand{\newblock}{\relax}
\providecommand{\bibinfo}[2]{#2}
\providecommand{\BIBentrySTDinterwordspacing}{\spaceskip=0pt\relax}
\providecommand{\BIBentryALTinterwordstretchfactor}{4}
\providecommand{\BIBentryALTinterwordspacing}{\spaceskip=\fontdimen2\font plus
\BIBentryALTinterwordstretchfactor\fontdimen3\font minus \fontdimen4\font\relax}
\providecommand{\BIBforeignlanguage}[2]{{%
\expandafter\ifx\csname l@#1\endcsname\relax
\typeout{** WARNING: IEEEtran.bst: No hyphenation pattern has been}%
\typeout{** loaded for the language `#1'. Using the pattern for}%
\typeout{** the default language instead.}%
\else
\language=\csname l@#1\endcsname
\fi
#2}}
\providecommand{\BIBdecl}{\relax}
\BIBdecl

\bibitem{keskar2019ctrl}
N.~S. Keskar, B.~McCann, L.~R. Varshney, C.~Xiong, and R.~Socher, ``Ctrl: A conditional transformer language model for controllable generation,'' \emph{arXiv preprint arXiv:1909.05858}, 2019.

\bibitem{Dathathri2020Plug}
\BIBentryALTinterwordspacing
S.~Dathathri, A.~Madotto, J.~Lan, J.~Hung, E.~Frank, P.~Molino, J.~Yosinski, and R.~Liu, ``Plug and play language models: A simple approach to controlled text generation,'' in \emph{International Conference on Learning Representations}, 2020. [Online]. Available: \url{https://openreview.net/forum?id=H1edEyBKDS}
\BIBentrySTDinterwordspacing

\bibitem{qian2022controllable}
J.~Qian, L.~Dong, Y.~Shen, F.~Wei, and W.~Chen, ``Controllable natural language generation with contrastive prefixes,'' \emph{arXiv preprint arXiv:2202.13257}, 2022.

\bibitem{ghosh2017affect}
S.~Ghosh, M.~Chollet, E.~Laksana, L.-P. Morency, and S.~Scherer, ``Affect-lm: A neural language model for customizable affective text generation,'' in \emph{Proceedings of the 55th Annual Meeting of the Association for Computational Linguistics (Volume 1: Long Papers)}, 2017, pp. 634--642.

\bibitem{li2020hierarchical}
Y.~Li, R.~Zhang, W.~Li, and Z.~Cao, ``Hierarchical prediction and adversarial learning for conditional response generation,'' \emph{IEEE Transactions on Knowledge and Data Engineering}, vol.~34, no.~1, pp. 314--327, 2020.

\bibitem{xing2017topic}
C.~Xing, W.~Wu, Y.~Wu, J.~Liu, Y.~Huang, M.~Zhou, and W.-Y. Ma, ``Topic aware neural response generation,'' in \emph{Proceedings of the AAAI conference on artificial intelligence}, vol.~31, no.~1, 2017.

\bibitem{liao2020topic}
L.~Liao, R.~Takanobu, Y.~Ma, X.~Yang, M.~Huang, and T.-S. Chua, ``Topic-guided conversational recommender in multiple domains,'' \emph{IEEE Transactions on Knowledge and Data Engineering}, vol.~34, no.~5, pp. 2485--2496, 2020.

\bibitem{liu2022graph}
Z.~Liu, D.~Zhou, H.~Liu, H.~Wang, Z.-Y. Niu, H.~Wu, W.~Che, T.~Liu, and H.~Xiong, ``Graph-grounded goal planning for conversational recommendation,'' \emph{IEEE Transactions on Knowledge and Data Engineering}, vol.~35, no.~5, pp. 4923--4939, 2022.

\bibitem{NEURIPS2022_871cae8f}
\BIBentryALTinterwordspacing
Y.~Su, T.~Lan, Y.~Wang, D.~Yogatama, L.~Kong, and N.~Collier, ``A contrastive framework for neural text generation,'' in \emph{Advances in Neural Information Processing Systems}, S.~Koyejo, S.~Mohamed, A.~Agarwal, D.~Belgrave, K.~Cho, and A.~Oh, Eds., vol.~35.\hskip 1em plus 0.5em minus 0.4em\relax Curran Associates, Inc., 2022, pp. 21\,548--21\,561. [Online]. Available: \url{https://proceedings.neurips.cc/paper_files/paper/2022/file/871cae8f599cb8bbfcb0f58fe1af95ad-Paper-Conference.pdf}
\BIBentrySTDinterwordspacing

\bibitem{su2022language}
Y.~Su, T.~Lan, Y.~Liu, F.~Liu, D.~Yogatama, Y.~Wang, L.~Kong, and N.~Collier, ``Language models can see: Plugging visual controls in text generation,'' 2022.

\bibitem{lin-etal-2020-commongen}
\BIBentryALTinterwordspacing
B.~Y. Lin, W.~Zhou, M.~Shen, P.~Zhou, C.~Bhagavatula, Y.~Choi, and X.~Ren, ``{C}ommon{G}en: A constrained text generation challenge for generative commonsense reasoning,'' in \emph{Findings of the Association for Computational Linguistics: EMNLP 2020}.\hskip 1em plus 0.5em minus 0.4em\relax Online: Association for Computational Linguistics, Nov. 2020, pp. 1823--1840. [Online]. Available: \url{https://www.aclweb.org/anthology/2020.findings-emnlp.165}
\BIBentrySTDinterwordspacing

\bibitem{li2023automatic}
Y.~Li, S.~Huang, X.~Zhang, Q.~Zhou, Y.~Li, R.~Liu, Y.~Cao, H.-T. Zheng, and Y.~Shen, ``Automatic context pattern generation for entity set expansion,'' \emph{IEEE Transactions on Knowledge and Data Engineering}, 2023.

\bibitem{lan2022complex}
Y.~Lan, G.~He, J.~Jiang, J.~Jiang, W.~X. Zhao, and J.-R. Wen, ``Complex knowledge base question answering: A survey,'' \emph{IEEE Transactions on Knowledge and Data Engineering}, 2022.

\bibitem{zhang2018personalizing}
S.~{Zhang}, E.~{Dinan}, J.~{Urbanek}, A.~{Szlam}, D.~{Kiela}, and J.~{Weston}, ``Personalizing dialogue agents: I have a dog, do you have pets too?'' in \emph{Proceedings of the 56th Annual Meeting of the Association for Computational Linguistics (Volume 1: Long Papers)}, vol.~1, 2018, pp. 2204--2213.

\bibitem{fan-etal-2020-enhanced}
\BIBentryALTinterwordspacing
Z.~Fan, Y.~Gong, Z.~Wei, S.~Wang, Y.~Huang, J.~Jiao, X.~Huang, N.~Duan, and R.~Zhang, ``An enhanced knowledge injection model for commonsense generation,'' in \emph{Proceedings of the 28th International Conference on Computational Linguistics}.\hskip 1em plus 0.5em minus 0.4em\relax Barcelona, Spain (Online): International Committee on Computational Linguistics, Dec. 2020, pp. 2014--2025. [Online]. Available: \url{https://aclanthology.org/2020.coling-main.182}
\BIBentrySTDinterwordspacing

\bibitem{majumder-etal-2020-like}
\BIBentryALTinterwordspacing
B.~P. Majumder, H.~Jhamtani, T.~Berg-Kirkpatrick, and J.~McAuley, ``Like hiking? you probably enjoy nature: Persona-grounded dialog with commonsense expansions,'' in \emph{Proceedings of the 2020 Conference on Empirical Methods in Natural Language Processing (EMNLP)}.\hskip 1em plus 0.5em minus 0.4em\relax Online: Association for Computational Linguistics, Nov. 2020, pp. 9194--9206. [Online]. Available: \url{https://aclanthology.org/2020.emnlp-main.739}
\BIBentrySTDinterwordspacing

\bibitem{wang2021neural}
Y.~Wang, C.~Xu, H.~Hu, C.~Tao, S.~Wan, M.~Dras, M.~Johnson, and D.~Jiang, ``Neural rule-execution tracking machine for transformer-based text generation,'' \emph{Advances in Neural Information Processing Systems}, vol.~34, pp. 16\,938--16\,950, 2021.

\bibitem{liu2021kg}
Y.~Liu, Y.~Wan, L.~He, H.~Peng, and S.~Y. Philip, ``Kg-bart: Knowledge graph-augmented bart for generative commonsense reasoning,'' in \emph{Proceedings of the AAAI Conference on Artificial Intelligence}, vol.~35, no.~7, 2021, pp. 6418--6425.

\bibitem{carlsson2022fine}
F.~Carlsson, J.~{\"O}hman, F.~Liu, S.~Verlinden, J.~Nivre, and M.~Sahlgren, ``Fine-grained controllable text generation using non-residual prompting,'' in \emph{Proceedings of the 60th Annual Meeting of the Association for Computational Linguistics (Volume 1: Long Papers)}, 2022, pp. 6837--6857.

\bibitem{liu-etal-2020-impress}
\BIBentryALTinterwordspacing
Q.~Liu, Y.~Chen, B.~Chen, J.-G. Lou, Z.~Chen, B.~Zhou, and D.~Zhang, ``You impress me: Dialogue generation via mutual persona perception,'' in \emph{Proceedings of the 58th Annual Meeting of the Association for Computational Linguistics}.\hskip 1em plus 0.5em minus 0.4em\relax Online: Association for Computational Linguistics, Jul. 2020, pp. 1417--1427. [Online]. Available: \url{https://aclanthology.org/2020.acl-main.131}
\BIBentrySTDinterwordspacing

\bibitem{xu2022cosplay}
C.~Xu, P.~Li, W.~Wang, H.~Yang, S.~Wang, and C.~Xiao, ``Cosplay: Concept set guided personalized dialogue generation across both party personas,'' in \emph{Proceedings of the 45th International ACM SIGIR Conference on Research and Development in Information Retrieval}, 2022, pp. 201--211.

\bibitem{wang2021contextualized}
P.~Wang, J.~Zamora, J.~Liu, F.~Ilievski, M.~Chen, and X.~Ren, ``Contextualized scene imagination for generative commonsense reasoning,'' in \emph{International Conference on Learning Representations (ICLR)}, 2022.

\bibitem{yang-etal-2023-bridging}
\BIBentryALTinterwordspacing
H.~Yang, Y.~Wang, P.~Li, W.~Bi, W.~Lam, and C.~Xu, ``Bridging the gap between pre-training and fine-tuning for commonsense generation,'' in \emph{Findings of the Association for Computational Linguistics: EACL 2023}, A.~Vlachos and I.~Augenstein, Eds.\hskip 1em plus 0.5em minus 0.4em\relax Dubrovnik, Croatia: Association for Computational Linguistics, May 2023, pp. 376--383. [Online]. Available: \url{https://aclanthology.org/2023.findings-eacl.28}
\BIBentrySTDinterwordspacing

\bibitem{wang-etal-2021-mention}
\BIBentryALTinterwordspacing
Y.~Wang, I.~Wood, S.~Wan, M.~Dras, and M.~Johnson, ``Mention flags ({MF}): Constraining transformer-based text generators,'' in \emph{Proceedings of the 59th Annual Meeting of the Association for Computational Linguistics and the 11th International Joint Conference on Natural Language Processing (Volume 1: Long Papers)}, C.~Zong, F.~Xia, W.~Li, and R.~Navigli, Eds.\hskip 1em plus 0.5em minus 0.4em\relax Online: Association for Computational Linguistics, Aug. 2021, pp. 103--113. [Online]. Available: \url{https://aclanthology.org/2021.acl-long.9}
\BIBentrySTDinterwordspacing

\bibitem{anderson-etal-2017-guided}
\BIBentryALTinterwordspacing
P.~Anderson, B.~Fernando, M.~Johnson, and S.~Gould, ``Guided open vocabulary image captioning with constrained beam search,'' in \emph{Proceedings of the 2017 Conference on Empirical Methods in Natural Language Processing}.\hskip 1em plus 0.5em minus 0.4em\relax Copenhagen, Denmark: Association for Computational Linguistics, Sep. 2017, pp. 936--945. [Online]. Available: \url{https://www.aclweb.org/anthology/D17-1098}
\BIBentrySTDinterwordspacing

\bibitem{hokamp-liu-2017-lexically}
\BIBentryALTinterwordspacing
C.~Hokamp and Q.~Liu, ``Lexically constrained decoding for sequence generation using grid beam search,'' in \emph{Proceedings of the 55th Annual Meeting of the Association for Computational Linguistics (Volume 1: Long Papers)}.\hskip 1em plus 0.5em minus 0.4em\relax Vancouver, Canada: Association for Computational Linguistics, Jul. 2017, pp. 1535--1546. [Online]. Available: \url{https://www.aclweb.org/anthology/P17-1141}
\BIBentrySTDinterwordspacing

\bibitem{post-vilar-2018-fast}
\BIBentryALTinterwordspacing
M.~Post and D.~Vilar, ``Fast lexically constrained decoding with dynamic beam allocation for neural machine translation,'' in \emph{Proceedings of the 2018 Conference of the North {A}merican Chapter of the Association for Computational Linguistics: Human Language Technologies, Volume 1 (Long Papers)}.\hskip 1em plus 0.5em minus 0.4em\relax New Orleans, Louisiana: Association for Computational Linguistics, Jun. 2018, pp. 1314--1324. [Online]. Available: \url{https://www.aclweb.org/anthology/N18-1119}
\BIBentrySTDinterwordspacing

\bibitem{pascual-etal-2021-plug-play}
\BIBentryALTinterwordspacing
D.~Pascual, B.~Egressy, C.~Meister, R.~Cotterell, and R.~Wattenhofer, ``A plug-and-play method for controlled text generation,'' in \emph{Findings of the Association for Computational Linguistics: EMNLP 2021}.\hskip 1em plus 0.5em minus 0.4em\relax Punta Cana, Dominican Republic: Association for Computational Linguistics, Nov. 2021, pp. 3973--3997. [Online]. Available: \url{https://aclanthology.org/2021.findings-emnlp.334}
\BIBentrySTDinterwordspacing

\bibitem{lu-etal-2021-neurologic}
\BIBentryALTinterwordspacing
X.~Lu, P.~West, R.~Zellers, R.~Le~Bras, C.~Bhagavatula, and Y.~Choi, ``{N}euro{L}ogic decoding: (un)supervised neural text generation with predicate logic constraints,'' in \emph{Proceedings of the 2021 Conference of the North American Chapter of the Association for Computational Linguistics: Human Language Technologies}.\hskip 1em plus 0.5em minus 0.4em\relax Online: Association for Computational Linguistics, Jun. 2021, pp. 4288--4299. [Online]. Available: \url{https://aclanthology.org/2021.naacl-main.339}
\BIBentrySTDinterwordspacing

\bibitem{qin2022cold}
L.~Qin, S.~Welleck, D.~Khashabi, and Y.~Choi, ``Cold decoding: Energy-based constrained text generation with langevin dynamics,'' \emph{Advances in Neural Information Processing Systems}, vol.~35, pp. 9538--9551, 2022.

\bibitem{daniel2017thinking}
K.~Daniel, \emph{Thinking, fast and slow}, 2017.

\bibitem{ji-etal-2022-controlling}
\BIBentryALTinterwordspacing
J.~Ji, Y.~Kim, J.~Glass, and T.~He, ``Controlling the focus of pretrained language generation models,'' in \emph{Findings of the Association for Computational Linguistics: ACL 2022}.\hskip 1em plus 0.5em minus 0.4em\relax Dublin, Ireland: Association for Computational Linguistics, May 2022, pp. 3291--3306. [Online]. Available: \url{https://aclanthology.org/2022.findings-acl.260}
\BIBentrySTDinterwordspacing

\bibitem{dong-etal-2021-fly}
\BIBentryALTinterwordspacing
Y.~Dong, C.~Bhagavatula, X.~Lu, J.~D. Hwang, A.~Bosselut, J.~C.~K. Cheung, and Y.~Choi, ``On-the-fly attention modulation for neural generation,'' in \emph{Findings of the Association for Computational Linguistics: ACL-IJCNLP 2021}.\hskip 1em plus 0.5em minus 0.4em\relax Online: Association for Computational Linguistics, Aug. 2021, pp. 1261--1274. [Online]. Available: \url{https://aclanthology.org/2021.findings-acl.107}
\BIBentrySTDinterwordspacing

\bibitem{bach2015hinge}
S.~H. Bach, M.~Broecheler, B.~Huang, and L.~Getoor, ``Hinge-loss {M}arkov random fields and probabilistic soft logic,'' \emph{arXiv preprint arXiv:1505.04406}, 2015.

\bibitem{foulds2015latent}
J.~Foulds, S.~Kumar, and L.~Getoor, ``Latent topic networks: A versatile probabilistic programming framework for topic models,'' in \emph{Proc. of ICML}, 2015, pp. 777--786.

\bibitem{colmerauer1990introduction}
A.~Colmerauer, ``An introduction to prolog iii,'' \emph{Communications of the ACM}, vol.~33, no.~7, pp. 69--90, 1990.

\bibitem{speer2017conceptnet}
R.~Speer, J.~Chin, and C.~Havasi, ``Conceptnet 5.5: An open multilingual graph of general knowledge,'' in \emph{Thirty-first AAAI conference on artificial intelligence}, 2017.

\bibitem{vaswani2017attention}
A.~Vaswani, N.~Shazeer, N.~Parmar, J.~Uszkoreit, L.~Jones, A.~N. Gomez, {\L}.~Kaiser, and I.~Polosukhin, ``Attention is all you need,'' in \emph{Advances in neural information processing systems}, 2017, pp. 5998--6008.

\bibitem{liu2023summary}
Y.~Liu, T.~Han, S.~Ma, J.~Zhang, Y.~Yang, J.~Tian, H.~He, A.~Li, M.~He, Z.~Liu \emph{et~al.}, ``Summary of chatgpt-related research and perspective towards the future of large language models,'' \emph{Meta-Radiology}, p. 100017, 2023.

\bibitem{radford2018improving}
A.~Radford, K.~Narasimhan, T.~Salimans, and I.~Sutskever, ``Improving language understanding by generative pre-training,'' 2018.

\bibitem{radford2019language}
A.~Radford, J.~Wu, R.~Child, D.~Luan, D.~Amodei, and I.~Sutskever, ``Language models are unsupervised multitask learners,'' 2019.

\bibitem{zhang-etal-2020-dialogpt}
\BIBentryALTinterwordspacing
Y.~Zhang, S.~Sun, M.~Galley, Y.-C. Chen, C.~Brockett, X.~Gao, J.~Gao, J.~Liu, and B.~Dolan, ``{DIALOGPT} : Large-scale generative pre-training for conversational response generation,'' in \emph{Proceedings of the 58th Annual Meeting of the Association for Computational Linguistics: System Demonstrations}.\hskip 1em plus 0.5em minus 0.4em\relax Online: Association for Computational Linguistics, Jul. 2020, pp. 270--278. [Online]. Available: \url{https://aclanthology.org/2020.acl-demos.30}
\BIBentrySTDinterwordspacing

\bibitem{fan2018hierarchical}
A.~Fan, M.~Lewis, and Y.~Dauphin, ``Hierarchical neural story generation,'' \emph{arXiv preprint arXiv:1805.04833}, 2018.

\bibitem{zhang2020language}
M.~Zhang, N.~Jiang, L.~Li, and Y.~Xue, ``Language generation via combinatorial constraint satisfaction: A tree search enhanced monte-carlo approach,'' in \emph{Findings of the Association for Computational Linguistics: EMNLP 2020}, 2020, pp. 1286--1298.

\bibitem{dong2021fly}
Y.~Dong, C.~Bhagavatula, X.~Lu, J.~D. Hwang, A.~Bosselut, J.~C. Cheung, and Y.~Choi, ``On-the-fly attention modulation for neural generation,'' \emph{Findings of the Association for Computational Linguistics: ACL-IJCNLP 2021}, 2021.

\bibitem{tian-etal-2024-sheng}
\BIBentryALTinterwordspacing
L.~Tian, M.~Ziao, Z.~Yanghao, X.~Chen, and M.~Xianling, ``A survey of automatic evaluation on the quality of generated text,'' in \emph{Proceedings of the 23rd Chinese National Conference on Computational Linguistics (Volume 2: Frontier Forum)}, X.~Zhao, Ed.\hskip 1em plus 0.5em minus 0.4em\relax Taiyuan, China: Chinese Information Processing Society of China, Jul. 2024, pp. 169--196. [Online]. Available: \url{https://aclanthology.org/2024.ccl-2.10/}
\BIBentrySTDinterwordspacing

\bibitem{papineni2002bleu}
K.~Papineni, S.~Roukos, T.~Ward, and W.-J. Zhu, ``Bleu: a method for automatic evaluation of machine translation,'' in \emph{Proceedings of the 40th annual meeting of the Association for Computational Linguistics}, 2002, pp. 311--318.

\bibitem{lin2004rouge}
C.-Y. Lin, ``Rouge: A package for automatic evaluation of summaries,'' in \emph{Text summarization branches out}, 2004, pp. 74--81.

\bibitem{banerjee2005meteor}
S.~Banerjee and A.~Lavie, ``Meteor: An automatic metric for mt evaluation with improved correlation with human judgments,'' in \emph{Proceedings of the acl workshop on intrinsic and extrinsic evaluation measures for machine translation and/or summarization}, 2005, pp. 65--72.

\bibitem{Vedantam_2015_CVPR}
R.~Vedantam, C.~Lawrence~Zitnick, and D.~Parikh, ``Cider: Consensus-based image description evaluation,'' in \emph{Proceedings of the IEEE Conference on Computer Vision and Pattern Recognition (CVPR)}, June 2015.

\bibitem{anderson2016spice}
P.~Anderson, B.~Fernando, M.~Johnson, and S.~Gould, ``Spice: Semantic propositional image caption evaluation,'' in \emph{European conference on computer vision}.\hskip 1em plus 0.5em minus 0.4em\relax Springer, 2016, pp. 382--398.

\bibitem{touvron2023llama}
H.~Touvron, L.~Martin, K.~Stone, P.~Albert, A.~Almahairi, Y.~Babaei, N.~Bashlykov, S.~Batra, P.~Bhargava, S.~Bhosale \emph{et~al.}, ``Llama 2: Open foundation and fine-tuned chat models,'' \emph{arXiv preprint arXiv:2307.09288}, 2023.

\bibitem{dubey2024llama}
A.~Dubey, A.~Jauhri, A.~Pandey, A.~Kadian, A.~Al-Dahle, A.~Letman, A.~Mathur, A.~Schelten, A.~Yang, A.~Fan \emph{et~al.}, ``The llama 3 herd of models,'' \emph{arXiv preprint arXiv:2407.21783}, 2024.

\bibitem{brown2020language}
T.~B. Brown, B.~Mann, N.~Ryder, M.~Subbiah, J.~Kaplan, P.~Dhariwal, A.~Neelakantan, P.~Shyam, G.~Sastry, A.~Askell \emph{et~al.}, ``Language models are few-shot learners,'' in \emph{Proceedings of the 34th International Conference on Neural Information Processing Systems}, 2020, pp. 1877--1901.

\bibitem{openai2023gpt}
R.~OpenAI, ``Gpt-4 technical report. arxiv 2303.08774,'' \emph{View in Article}, vol.~2, no.~5, 2023.

\bibitem{cao2022model}
Y.~Cao, W.~Bi, M.~Fang, S.~Shi, and D.~Tao, ``A model-agnostic data manipulation method for persona-based dialogue generation,'' \emph{arXiv preprint arXiv:2204.09867}, 2022.

\bibitem{bert-score}
\BIBentryALTinterwordspacing
T.~Zhang, V.~Kishore, F.~Wu, K.~Q. Weinberger, and Y.~Artzi, ``Bertscore: Evaluating text generation with bert,'' in \emph{International Conference on Learning Representations}, 2020. [Online]. Available: \url{https://openreview.net/forum?id=SkeHuCVFDr}
\BIBentrySTDinterwordspacing

\bibitem{he2021deberta}
\BIBentryALTinterwordspacing
P.~He, X.~Liu, J.~Gao, and W.~Chen, ``Deberta: Decoding-enhanced bert with disentangled attention,'' in \emph{International Conference on Learning Representations}, 2021. [Online]. Available: \url{https://openreview.net/forum?id=XPZIaotutsD}
\BIBentrySTDinterwordspacing

\bibitem{liu2019roberta}
Y.~Liu, M.~Ott, N.~Goyal, J.~Du, M.~Joshi, D.~Chen, O.~Levy, M.~Lewis, L.~Zettlemoyer, and V.~Stoyanov, ``Roberta: A robustly optimized bert pretraining approach,'' 2019.

\bibitem{welleck-etal-2019-dialogue}
\BIBentryALTinterwordspacing
S.~Welleck, J.~Weston, A.~Szlam, and K.~Cho, ``Dialogue natural language inference,'' in \emph{Proceedings of the 57th Annual Meeting of the Association for Computational Linguistics}.\hskip 1em plus 0.5em minus 0.4em\relax Florence, Italy: Association for Computational Linguistics, Jul. 2019, pp. 3731--3741. [Online]. Available: \url{https://www.aclweb.org/anthology/P19-1363}
\BIBentrySTDinterwordspacing

\bibitem{song-etal-2021-bob}
\BIBentryALTinterwordspacing
H.~Song, Y.~Wang, K.~Zhang, W.-N. Zhang, and T.~Liu, ``{B}o{B}: {BERT} over {BERT} for training persona-based dialogue models from limited personalized data,'' in \emph{Proceedings of the 59th Annual Meeting of the Association for Computational Linguistics and the 11th International Joint Conference on Natural Language Processing (Volume 1: Long Papers)}.\hskip 1em plus 0.5em minus 0.4em\relax Online: Association for Computational Linguistics, Aug. 2021, pp. 167--177. [Online]. Available: \url{https://aclanthology.org/2021.acl-long.14}
\BIBentrySTDinterwordspacing

\bibitem{madotto-etal-2019-personalizing}
\BIBentryALTinterwordspacing
A.~Madotto, Z.~Lin, C.-S. Wu, and P.~Fung, ``Personalizing dialogue agents via meta-learning,'' in \emph{Proceedings of the 57th Annual Meeting of the Association for Computational Linguistics}.\hskip 1em plus 0.5em minus 0.4em\relax Florence, Italy: Association for Computational Linguistics, Jul. 2019, pp. 5454--5459. [Online]. Available: \url{https://aclanthology.org/P19-1542}
\BIBentrySTDinterwordspacing

\bibitem{liu2020you}
Q.~Liu, Y.~Chen, B.~Chen, J.-G. Lou, Z.~Chen, B.~Zhou, and D.~Zhang, ``You impress me: Dialogue generation via mutual persona perception,'' in \emph{Proceedings of the 58th Annual Meeting of the Association for Computational Linguistics}, 2020, pp. 1417--1427.

\bibitem{susskind2021neuro}
Z.~Susskind, B.~Arden, L.~K. John, P.~Stockton, and E.~B. John, ``Neuro-symbolic ai: An emerging class of ai workloads and their characterization,'' \emph{arXiv preprint arXiv:2109.06133}, 2021.

\bibitem{li2020knowledge}
\BIBentryALTinterwordspacing
Q.~Li, P.~Li, Z.~Ren, P.~Ren, and Z.~Chen, ``Knowledge bridging for empathetic dialogue generation,'' 2020. [Online]. Available: \url{https://arxiv.org/abs/2009.09708}
\BIBentrySTDinterwordspacing

\bibitem{shen2024cbt}
H.~Shen, Z.~Li, M.~Yang, M.~Ni, Y.~Tao, Z.~Yu, W.~Zheng, C.~Xu, and B.~Hu, ``Are large language models possible to conduct cognitive behavioral therapy?'' in \emph{2024 IEEE International Conference on Bioinformatics and Biomedicine (BIBM)}.\hskip 1em plus 0.5em minus 0.4em\relax IEEE, 2024, pp. 3695--3700.

\bibitem{ji2020language}
H.~{Ji}, P.~{Ke}, S.~{Huang}, F.~{Wei}, X.~{Zhu}, and M.~{Huang}, ``Language generation with multi-hop reasoning on commonsense knowledge graph.'' in \emph{Proceedings of the 2020 Conference on Empirical Methods in Natural Language Processing (EMNLP)}, 2020, pp. 725--736.

\bibitem{zhou2021clinical}
Y.~Zhou, Y.~Yan, R.~Han, J.~H. Caufield, K.-W. Chang, Y.~Sun, P.~Ping, and W.~Wang, ``Clinical temporal relation extraction with probabilistic soft logic regularization and global inference,'' in \emph{Proceedings of the AAAI Conference on Artificial Intelligence}, vol.~35, no.~16, 2021, pp. 14\,647--14\,655.

\bibitem{hu-etal-2016-harnessing}
\BIBentryALTinterwordspacing
Z.~Hu, X.~Ma, Z.~Liu, E.~Hovy, and E.~Xing, ``Harnessing deep neural networks with logic rules,'' in \emph{Proceedings of the 54th Annual Meeting of the Association for Computational Linguistics (Volume 1: Long Papers)}, Berlin, Germany, Aug. 2016, pp. 2410--2420. [Online]. Available: \url{https://aclanthology.org/P16-1228}
\BIBentrySTDinterwordspacing

\bibitem{nye2021improving}
M.~Nye, M.~Tessler, J.~Tenenbaum, and B.~M. Lake, ``Improving coherence and consistency in neural sequence models with dual-system, neuro-symbolic reasoning,'' \emph{Advances in Neural Information Processing Systems}, vol.~34, pp. 25\,192--25\,204, 2021.

\bibitem{shu-etal-2021-logic}
\BIBentryALTinterwordspacing
C.~Shu, Y.~Zhang, X.~Dong, P.~Shi, T.~Yu, and R.~Zhang, ``Logic-consistency text generation from semantic parses,'' in \emph{Findings of the Association for Computational Linguistics: ACL-IJCNLP 2021}.\hskip 1em plus 0.5em minus 0.4em\relax Online: Association for Computational Linguistics, Aug. 2021, pp. 4414--4426. [Online]. Available: \url{https://aclanthology.org/2021.findings-acl.388}
\BIBentrySTDinterwordspacing

\bibitem{xu2021change}
C.~Xu, J.~Zhao, R.~Li, C.~Hu, and C.~Xiao, ``Change or not: A simple approach for plug and play language models on sentiment control,'' in \emph{Proceedings of the AAAI Conference on Artificial Intelligence}, vol.~35, no.~18, 2021, pp. 15\,935--15\,936.

\bibitem{wolf2019huggingface}
T.~Wolf, L.~Debut, V.~Sanh, J.~Chaumond, C.~Delangue, A.~Moi, P.~Cistac, T.~Rault, R.~Louf, M.~Funtowicz \emph{et~al.}, ``Huggingface's transformers: State-of-the-art natural language processing,'' \emph{arXiv preprint arXiv:1910.03771}, 2019.

\bibitem{paszke2019pytorch}
A.~Paszke, S.~Gross, F.~Massa, A.~Lerer, J.~Bradbury, G.~Chanan, T.~Killeen, Z.~Lin, N.~Gimelshein, L.~Antiga \emph{et~al.}, ``Pytorch: An imperative style, high-performance deep learning library,'' \emph{Advances in neural information processing systems}, vol.~32, pp. 8026--8037, 2019.

\end{thebibliography}

\end{document}